\documentclass{llncs}
\usepackage{graphics} 
\usepackage{times} 
\usepackage{amsmath} 
\usepackage{amssymb}  
\usepackage{epsfig}  
\usepackage{color}

\begin{document}

\title{A Logic of Agent Organizations}
\author{Virginia Dignum\inst{1} \and Frank Dignum\inst{2}}
\institute{
    Delft University of Technology - Dept. Technology, Policy and Management\\
    The Netherlands\\
    email: m.v.dignum@tudelft.nl \and
    Utrecht University - Institute of Information and Computing Sciences\\
    The Netherlands\\
	email: dignum@cs.uu.nl }
\maketitle

\begin{abstract}
Organization concepts and models are increasingly being adopted for the design and specification of
multi-agent systems. Agent organizations can be seen as mechanisms of social order, created to achieve global (or organizational) objectives by more or less autonomous agents. In order to develop a theory on the relation between organizational structures, organizational objectives and the actions of agents fulfilling roles in the organization a theoretical framework is needed to describe organizational structures and actions of (groups of) agents. Current logical formalisms focus on specific aspects of organizations (e.g. power, delegation, agent actions, or normative issues) but a framework that integrates and relates different aspects is missing. Given the amount of aspects involved and the subsequent complexity of a formalism encompassing them all, it is difficult to realize. In this paper, a first step is taken to solve this problem. We present a generic formal model  that enables to specify and relate the main concepts of an organization (including, activity, structure, environment and others) so that organizations can be analyzed at a high level of abstraction. However, for some aspects we use a simplified model in order to avoid the complexity of combining many different types of (modal) operators.
\end{abstract}

\section{Introduction}\label{sect:intro}
The growing complexity of (information) systems, characterized by distribution,
heterogeneity, openness and dynamicity has lead to an increasing interest in
organizational concepts by MAS researchers. As in human organizations, the
specification of an explicit organization for a MAS helps coordinating the
agents' autonomous behavior \cite{hubner:ooop:06,barber:01}. Traditional design
of MAS is agent-centered, that is, mainly concerned with the representation of
an agent's internal knowledge or behavior. Recent research has pointed out that
an agent-centered perspective on MAS has several problems. According to
Jennings \cite{jennings:00}: \emph{``the patterns and the outcomes of the
interactions are inherently unpredictable, and predicting the behavior of the
overall system based on its constituent components is extremely difficult
(sometimes impossible) because of the high likelihood of emergent (and
unwanted) behavior"}.

A possible approach to organization in MAS is to look at how human
organizations are organized and try to apply results from many decades of
research in Organizational Theory (OT) to agent systems. Researchers in computer science and artificial intelligence are increasingly adopting concepts from the OT field to design more efficient and flexible distributed systems
(\cite{so-durfee:98,cohen:86,fox:81}). In some of this work organizational structures are used in the design phase of the MAS and agents built as a kind of refinement of these structures. Thus agents and organizations are not independent entities. In earlier work, we have propagated to view organizations as separate entities from agents that also exist after the MAS is created. This is especially important in cases of open agent systems, where agents created by third parties can enter and leave. Taking this point of view, agents are not just fulfilling roles of an organization and following the patterns of interaction set out by the organizational structures. They have an independent existence (like in human organizations people have an existence independent from the organization). Following this approach, it becomes very important to describe the exact relation between the organizational entities on the one hand and the agents on the other hand.\\
In earlier work, we have taken an OT view to define organizations as a set of entities regulated by mechanisms of social
order and created by more or less autonomous actors to achieve common objectives
\cite{virginia:phd:04}. However, due to its nature, OT research tends to be not very formal from a computational perspective, 
which makes it difficult to move from its use as a concept or paradigm towards
applying social and organizational concepts for the formalization of MAS social
concepts. On the other hand, existing formal models are often limited to a specific domain
and are difficult to validate \cite{harrison:07}. That is, in general, logic lends itself better for the detailed analysis of one particular aspect of reality.


The above considerations provide two different motivations for this work. On the one hand, the need for a
formal representation of organizations, with their environment, objectives and agents in a way that
enables to analyze their partial contributions to the performance of the organization in a changing
environment. On the other hand, the need for such a model to be realistic enough to incorporate the more `pragmatic'
considerations faced by real organizations. Most existing formal models lack this realism, e.g. either by
ignoring temporal issues, or by taking a very restrictive view on the controllability of agents, or by
assuming complete control and knowledge within the system (cf.
\cite{hoek-wooldridge:05,santos-jones-carmo:97}).

Much of our own work in the last decade, has also focused on providing detailed formalisms of a specific aspect of social aspects of agent systems (e.g. norms, responsibility, delegation, counts-as,...). However, to date there are no comprehensive models that integrate all these issues. One of the reasons for this lack, is that a complete model is necessarily too complex to be done at once. In this paper, we take the middle ground by incorporating many aspects that are required for realism, but sometimes take simple formalizations in order to keep the logic manageable, where we refer to other work that could be used to extend some of the aspects wherever possible. The main goal of this article is rather an analysis of organizational
structures and their function on a high level, that enables to represent the foundations of agent organizations without considering all the details involved. This framework includes all main concepts involved in organizational modeling, and describes a kind of placeholders where later on more complete theories for detailed concepts can be
inserted. In a sense, the work presented in this paper simplifies many of the concepts for which powerful logics have been developed, but this simplification enables the integration of all aspects in one formalism.
E.g. we will reduce the concept of \emph{responsibility} to the concept of (organizational)
\emph{initiative}. Initiative can be seen as part of responsibility. It indicates that a certain role in the
organization is made responsible to perform a task and the initiative to do something about it lays with that
role.

Organizational analysis and planning aim to capture the functioning and development of organizations that account for the ways in which organizations respond to and bring about changes, and to design an organization's structure accordingly to bring about greater efficiency.
Different approaches to organizational analysis and planning have evolved from the rational view propagated by Taylor at the beginning of the 20th century \cite{taylor:1911}, to cognitive models of organizations, and include also holistic and sociotechnical system views. These views bring forward the need to identify structures, components, objectives and environment as separate but interelated aspects of organization \cite{helms:06}. 
This implies that formal models for organizations must meet the following requirements:
\begin{enumerate}
    \item represent notions of ability and activity of agents, without requiring knowledge about the
        specific actions available to a specific agent (open environments),
    \item accept limitedness of agent capability,
    \item represent the ability and activity of a group of agents,
    \item deal with temporal issues, especially the fact that activity takes time,
    \item represent the concept of 'being responsible' for the achievement of a given state of affairs,
    \item represent global (organizational) objectives and their dependency on agents' activity, by  relating
        activity and organizational structure,
    \item represent organizations in terms of organizational roles \label{req:role}
    \item relate roles and agents (role enacting agents) \label{req:rea}
    \item \emph{reason about organizational workflows and task dependencies (including partial ordering of tasks, parallelism, incompatibilities),}\label{req:workflow}
    \item \emph{represent organizational dynamics (evolution of the organization over time, changes of the
        agent population),}\label{req:dynamics}
    \item \emph{deal with resource limitedness and the dependency of activity on resources (e.g. costs),}
    \item \emph{deal with normative issues (representation of boundaries for action and the violation
        thereof)}\label{req:norm}
\end{enumerate}

In the remainder of this paper, we will describe a thoroughly revised version of the Logic for Agent
Organizations (LAO) introduced in \cite{dignum:omas:chapter:09} that will be increasingly extended to include
most of these requirements. The introduction of LAO, based on CTL* is done incrementally starting with the first
requirements and extend it successively to incorporate more complex requirements. The formalism presented in
this paper does not yet include requirements \ref{req:workflow} to \ref{req:norm}.
These issues have been dealt with in some of our other work, e.g. we dealt with normative issues extensively in e.g. \cite{grossi:coin-ubi:06}.
It is a matter for future work to connect the work reported in this article to that other
research. In the same way, more in-depth characterizations of other agency and social concepts can be given
by incorporating existing work in, e.g., power and delegation \cite{jonessergot:96,castelfranchi:95}. Where
appropriate we will indicate the place where these more elaborate theories can be substituted into the
framework for the basic relations that we adopted for the present paper.

This article is organized as follows. Section \ref{sect:organization} gives the background for this research and discusses related work. The following sections build the formal logical framework of LAO for
organizations and reorganization. In Section \ref{sect:lao-agt} we define the basis for organizations through
abilities, activities and related concepts (which we all together call \emph{achievement} modalities) of
agents and groups. In Section \ref{sect:lao-org}, the formal model for organizations including
structural and interaction properties is presented. Section \ref{sect:analysis} proposes an axiomatization for LAO.
Section \ref{sec:application} shows how LAO can be applied to provide a formalization of different
organization types. Finally, Section \ref{sect:conclusions} presents our conclusions and directions for
future work.

\section{Organization Theory}\label{sect:organization}

We use concepts from Organization Theory (OT) as basis for this work. Even though it can be said that OT lacks formality, its concepts have been studied and used in practice for many years with successful results. OT defines
\emph{Organization} as an entity that allows and supports an individual (be it a person, a computer system,
or an institution) to recognize its role, and the roles of others, in accomplishing collective goals. 
Furthermore, OT recognizes that
organizations are instruments of purpose, that is they are seen as coordinated by intentions and objectives
\cite{march:97}. 

Furthermore, organizations are assumed to have both a structural and a strategic component
\cite{donaldson:01}. These two components are linked in the following way. The organizational strategy
influences contingencies such as size, innovation, diversification and geographic distribution of the
organization. These characteristics result in different coordination mechanisms. E.g. if the whole
organization resides at the same physical location informal communication can be used to react quick to any
situation, but this is more difficult when the organization is spread all over the globe. Coordination
mechanisms are also dependent on task uncertainty and task dependency. Different organizational structures
are more appropriate for different types of tasks. E.g. factories usually have more hierarchical organization
structures than consultancy companies. A change in the strategy of the organization might change both the
tasks and the contingency factors and thus will have consequences for the organizational structure.

OT has for many decades investigated the issue of organizational structure. Organizational structure has
essentially two purposes \cite{duncan:79}: (1) it facilitates the flow of information within the
organization in order to reduce the uncertainty of decision making; and (2) it integrates organizational
behavior across the parts of the organization so that it is coordinated. Because the organizational structure
is defined over the roles within an organization and not the individual agents it also lends stability over
time to the organization. The organizational structure raises two challenges: division of labor and
coordination \cite{mintzberg:93}. The design of organizational structure determines the allocation of
resources and people to specified tasks or purposes (through the roles those people fulfill), and the
coordination of these resources to achieve organizational objectives \cite{galbraith:77}. Ideally, the
organization is designed to fit its environment and to provide the information and coordination needed for
its strategic objectives.

\subsection{Elements of organization}\label{section:orgelements}
Inspired by \cite{so-durfee:98}, we classify components of organizations into
three broad classes. The first are \emph{(task) environmental factors}, which
are the components and features of the task (such as size, time constraints,
uncertainty). The second are the \emph{structural factors}, which are the
components and features of the organization (such as roles, dependencies,
constraints, norms and regulations). The third class of factors are \emph{agent
factors}, which are the characteristics of the individual agents concerning
task capability, intelligence (including decision making and reasoning
capabilities), social awareness, etc. These three classes of factors jointly
determine the performance of the organization. To sum up, the three main issues
that must be represented in any model aimed at understanding or specifying
organizational performance or behavior are:
\begin{enumerate}
    \item Environment/world: this is the space in which
    organizations exist. This space is not completely controllable
    by the agents and results of agent activity are not
    guaranteed.
    \item Structure: describes the roles and relationships holding in the organization,
    and the strategy indicating the intentions of the organization
    in terms of objective(s).
    \item Agents: are entities with the capability to act, that is to control
    the state of some element in the environment. Agents are
    defined by their capabilities.
\end{enumerate}

In accordance to our previous work (e.g. \cite{virginia:phd:04}, organizational roles, like agents, are first class concepts in our model that provide an explicit separation between organization and the agents that act on it, and enable to separate organizational objectives from agent goals. We therefore assume that agents do not necessarily share the objectives of the organization.

\subsection{A Scenario for Organization}\label{section:scenario}
The study of organizations is particularly interesting in understanding and
managing the congruence of organizations with their environment. That is, why
is it that some organization thrive under given conditions, while others fail?
Due to their characteristics of autonomy, social conformance, flexibility,
collaborative problem solving, mobility, and distributed architecture, agent
based modelling and/or simulation are widely used in the area of organization
analysis. In particular, agent systems, have been advocated for
modelling supply-chain applications \cite{fox:00,nissen:01}. 
In this section we present a scenario to illustrate the formal modelling of
organizations proposed in this paper. The scenario is based on the work
described by Pelletier et al. \cite{pelletier:05} on the analysis of
reorganization of the Dutch gas pipeline transport market. Economic reforms
from the last decade have impacted a large reorganization of supply chain
processes, such as the opening of traditional monopolist markets to competition. 

The classic model for the gas market was a simple linear value chain. In this
situation organization is characterized by a fairly stable environment, where
parties and interactions are known and tasks are mostly fixed. In this
situation, hierarchies are proven to be efficient and robust
\cite{mintzberg:93}. The stakeholders in this situation are the state acting as
a (local) \emph{monopolist}, who organizes and directs the transmission of gas
from producer to consumer by controlling and directing other parties involved,
namely the \emph{trader} (responsible for exchanging gas from wellhead producer
to a local geographic area), the \emph{shipper} (responsible for transport of
high-pressure gas from origin to destination) and the \emph{local transport
manager} (responsible for the overall capacity and flow of gas).

%
%
Following the political decision on the liberalization of the gas market in the
Netherlands, the role of the monopolist disappears and the possibility for the
other partners to directly contract each other and for multiple parties to
enter the market is created. The environment became much more dynamic, with
different parties having the possibility to create and manage supply chains and
new services for the costumer. When the monopolist role is removed, the
existing hierarchical structure collapses and steps must be taken to assure
that the overall organization will still be able to achieve its objectives.
Typically, such situation will demand extra capabilities from the parties
involved in order to coordinate their activities without control by the
monopolist.

\section{Formal Core of  Agent Organization}\label{sect:lao-agt}
In this section we present the basis of the Logic of Agent Organization (LAO) that includes environment,
agents and organization structure. We start by formally defining the environment as a set of temporally
ordered set of states of affairs, or worlds. Secondly, we define the agents (and groups) in terms of their
capabilities, abilities and activity possibilities. Finally, we will formally define organizational structure
and activity, based on responsibilities and objectives.

\subsection{Environment}
In OT, the environment is commonly defined as the forces outside the organization that can impact it. The
environment changes over time is not fully controlled by the organizations and individuals that populate it. For a formal representation of
the environment, we use Kripke semantics to describe the environment and its changes. In the following, a
world describes an actual state of the environment, and transitions indicate the effects of possible changes
(the opportunities and threats in terms of organizational theory concepts). The set of propositional
variables $\Phi$ describes the vocabulary (or ontology) of the organization domain.

We use the well-known branching time temporal logic CTL* \cite{emerson:90} to describe  basic organization
and environment concepts. For a set $\Phi$ of propositional variables, the language, $\mathcal{L}$ consists
of two subsets: $\mathcal{L}_s$ the set of state formulae and $\mathcal{L}_p$ the set of path formulae,
defined as follows, where we use the usual CTL* notation: $\mathcal{A}$: all, $\mathcal{E}$: there is,
$\mathcal{X}$: next, $\mathcal{F}$: future, $\mathcal{G}$: always in future, and $\mathcal{U}$: until.
\begin{enumerate}
    \item $true, false \in \mathcal{L}_s$
    \item $p \in \Phi \Rightarrow p \in \mathcal{L}_s$
    \item $\varphi,\psi \in \mathcal{L}_s \Rightarrow \neg\varphi, (\varphi \vee \psi) \in \mathcal{L}_s$
    \item $\varphi \in \mathcal{L}_p \Rightarrow \mathcal{A}\varphi, \mathcal{E}\varphi \in
        \mathcal{L}_s$
    \item $\varphi \in \mathcal{L}_s \Rightarrow \varphi \in \mathcal{L}_p$
    \item $\varphi,\psi \in \mathcal{L}_p \Rightarrow  \neg\varphi, (\varphi \vee \psi),
        \mathcal{X}\varphi, \mathcal{F}\varphi, \mathcal{G}\varphi, \psi\mathcal{U}\varphi \in
        \mathcal{L}_p$
\end{enumerate}
A simple semantic structure over which formulae of $\mathcal{L}$ are interpreted is a tuple
\begin{equation}\label{eq:languageL}
  M = (W, Rt, \pi)
\end{equation}
where:
\begin{itemize}
    \item $W$ is a non empty set of states, 
    \item $Rt$ is a partial ordered set of (temporal) transitions between two elements of $W$, $Rt: W \times W$,
    \item $\pi$ is a valuation function which associates each $w \in W$ with the set of atomic propositions from $\Phi$ that are
    true in that world, $\pi: W \rightarrow 2^{\Phi}$ 
\end{itemize}
Each world $w \in W$ describes the propositions of $\Phi$ that are true in that world, and, each proposition
in $\Phi$ corresponds to a set of worlds where it is true. A transition between worlds represents an update
of the truth value of (some) propositions in $\Phi$. The semantics of CTL* \cite{emerson:90}\texttt{}
distinguish between path and state formulae. A state formula is interpreted wrt a state $w \in W$ and a path
formula is interpreted wrt a path through the branching time structure given by $Rt$. A path (or trace) in
$Rt$ is a (possibly infinite) sequence $(w_i, w_{i+1},...)$, where $w_i, w_{i+1},...\in W$ and $\forall i:
(w_i, w_{i+1}) \in Rt$. We use the convention that $rt = (w_0, w_1,...)$ denotes a path, and $rt(i)$ denotes
state $i$ in path $rt$. We write $M, w \models \varphi$ (resp. $M, rt \models \varphi$) to denote that state
formula $\varphi$ (resp. path formula $\varphi$) is true in structure $M$ at state $w$ (resp. path $rt$).
\begin{definition}[World Semantics]\label{def:worldsemantics}
The rules for the satisfaction relation $\models$ for state and path formulae are defined as:\\
\indent
\begin{tabular}{l}
    $M, w \models \top$\\
    $M, w \models p$ iff $p \in \pi(w)$, where $p\in \Phi$\\
    $M, w \models \neg\varphi$ iff not $M, w \models \varphi$\\
    $M, w \models \varphi \vee \psi$ iff $M, w \models \varphi$ or $M, w \models \psi$\\
    $M, w \models \mathcal{A}\varphi$ iff $\forall rt \in paths(W,Rt)$, if $rt(0) = w$ then $M, rt \models \varphi$\\
    $M, w \models \mathcal{E}\varphi$ iff $\exists rt \in paths(W,Rt)$, $rt(0) = w$ and $M, rt \models \varphi$\\
    $M, rt \models p$ iff $M, rt(0) \models p$, where $p\in \Phi$  \\
    $M, rt \models \neg\varphi$ iff not $M, rt \models \varphi$  \\
    $M, rt \models \varphi \vee \psi$ iff $M, rt \models \varphi$ or $M, rt \models \psi$\\
    $M, rt \models \mathcal{F}\varphi$ iff  $\exists i (M, rt(i)) \models \varphi$\\
    $M, rt \models \mathcal{G}\varphi$ iff  $\forall i (M, rt(i)) \models \varphi$\\
    $M, rt \models \mathcal{X}_p\varphi$ iff  $M, rt(1) \models \varphi$\\
    $M, rt \models \psi\mathcal{U}\varphi$ iff $\exists i$ such that $M, rt(i) \models \varphi$ and $\forall 0 \leq k < i, M, rt(k) \models \psi$\\
\end{tabular} \\
\end{definition}
Because there are two constructs that we will use very frequently we define some abbreviations for them:
\begin{itemize}
\item $\mathcal{X}\varphi \equiv \mathcal{AX}_p\varphi$
\item $\lozenge\varphi \equiv \mathcal{AF}\varphi$
\end{itemize}
This semantic structure enables the specification of the organization environment, with the affairs holding
in a state and the possible changes of state.

\subsection{Agents and groups}
The semantic structure described in the previous section does not consider the agents in the system.
Intuitively, the idea is that, in organizations, changes are for some part the result of the intervention of (specific) agents.
The language $\mathcal{L}_O$ for LAO is defined as an extension of the
language $\mathcal{L}$ described above with semantics given as an extension and slight modification to
CTL*. The semantics of $\mathcal{L}_O$ includes the rules listed in Definition \ref{def:worldsemantics}, and is extended
with rules for the other modalities described in the remainder of this section. 
Before we give a formal definition of $\mathcal{L}_O$ we first change the semantic structure over which
formulae of $\mathcal{L}_O$ will be interpreted and which will form the basis on which the new modalities can
be given an intuitive semantics.
\begin{equation}\label{eq:languageLAO}
  M_O = (\Psi, A, W, Rt, T, \pi, c)
\end{equation}
where:
\begin{itemize}
    \item $\Psi$ is a finite, non-empty set of \emph{domain facts}

    \item $W \subseteq 2^{\Psi}$ is a non-empty set of states. Each state indicates a set of domain facts that hold in that state.
    \item $Rt$ is as in (\ref{eq:languageL}),
    \item $\pi: \Psi \rightarrow \Phi$ is an interpretation function that associates an element of $\Phi$ with an element of $\Psi$.\\
        We overload the meaning of $\pi$ by also using it to associate sets of domain facts with sets of atomic propositions ($\pi: 2^{\Psi} \rightarrow 2^{\Phi}$). This allows us to use $\pi$ again as an evaluation function that associates with each world the set of atomic propositions from $\Phi$ that are true in that world.
    \item $A = \{a_1,...,a_n\}$ is a finite, non-empty set of \emph{agents},
    \item $c: A \times W \rightarrow 2^{\Psi}$ is a function that associates elements of $A$ and $W$ with the
    facts in $\Psi$
    \item $T: Rt \rightarrow 2^A$ is the set of agent labels on elements of $Rt$,
\end{itemize}
The first thing to notice is that we introduced a set of \emph{domain facts} next to the set of atomic
propositions. This is done for two reasons. First of all the capabilities of agents will be defined over the
domain facts. I.e. agents can change the truth values of atomic propositions that are linked to the domain
facts (and formulas derived from them). However, agents are not capable to control all possible atomic
propositions. 
Secondly, unlike the set of atomic propositions, the set of domain facts is per
definition finite. Although this limits the expressiveness of the logic, this restriction still maintains the requirement of realism for organization models as usually the set of facts considered important for an organization is captured in finite databases. This assumption simplifies a number of definitions later on (allowing us to use operations on finite sets and using universal quantification as just an abbreviation of enumeration of facts) and is not too restrictive for the domain facts, but would be for the set of all atomic propositions.

The set of worlds is built on the set of domain facts such that we can easily connect atomic propositions to
worlds using the domain facts.

This model also contains a set of agents and two relations that link the agents to the CTL* structure. The
first relation that links agents to the original CTL* structure is $c$. This relation indicates the
capabilities of agents in a particular state. If $p \in c(a,w)$ then agent $a$ has the capability to change
the truth value of $p$ when in state $w$. This relation thus forms the basis of what sort of things agents
are capable to influence.\\
The relation $T$ links sets of agents to particular transitions. These sets of
agents indicate the agents that influence the changes on that transition. That is, for a transition
$rt=(w,w') \in Rt$, $T(rt)$ indicates the set of agents that indeed contribute to the changes indicated by
that transition. Moreover, for each world $w\in W$ and each agent $a\in A$ we can indicate the set of
transitions starting from $w$ for which $a$ has influence.


\begin{definition}[Transition influence]\label{def:transitioninfluence}
For a model~\footnote{Note that in the remainder, we will drop the reference to the semantic model whenever
it is clear from the context.} $M_O = (\Psi, A, W, Rt, T, \pi, c)$, a world $w \in W$ and an agent $a \in A$,
the transition influence of $a$ in $w$, $T_{aw}$ is defined by: $T_{aw} = \{rt | rt \in Rt: \exists w': rt =
(w,w')$ and $a \in T(rt)\}$
\end{definition}

Note that, at any moment, unexpected change can also occur, which is not a result of the action of any of the
agents in that world (i.e. unlabeled or partially labeled transitions are possible).

\subsubsection{Agent Activity.}
The notions of agent capability and action have been widely discussed in MAS. In this section, we draw from
work in the area of the well known logical theory for agency and organized interaction introduced by
Kanger-Lindahl-P\"{o}rn, more specifically from the work of Santos et al. \cite{santos-jones-carmo:97} and
Governatori et al. \cite{governatori:03}. In short, their works assume that in realistic situations not all
capabilities are always conductive of successful action - one can attempt to achieve something but without
success. Three modal operators $E$, $G$ and $H$ are used to described these differences. The first one,
$E$, 
expresses direct and successful actions: i.e. a formula like $E_i\varphi$ means that the agent $i$
\emph{brings it about} that $\varphi$, that is, $\varphi$ is a necessary result of an action by $i$. The
second one, $G_i\varphi$ corresponds to indirect and successful actions, i.e., $G_i\varphi$ means that $i$
\emph{ensures} that $\varphi$, that is, $\varphi$ is a necessary result of an action of some agent following
an action by $i$. Finally, $H_i\varphi$ means that $i$ \emph{attempts} to make it the case that $\varphi$.
The idea is that $H$ is not necessarily successful. An axiomatic definition of $E, G$ and $H$ is given in
\cite{santos-jones-carmo:97,governatori:03}.

One of the main reasons to start from these abstract operators for actions and attempts is that they allow us
to conceive agents to be a kind of black boxes. We do not have to specify which precise actions and
procedures the agents actually have to execute in order to reach a certain situation. From an organizational perspective, agents are autonomous and have different decision mechanisms (and even architectures and/or implementation). Only their  outward behavior can be observed (i.e. agents are black boxes for the organization) and an attempt to perform an action can only be described if it is `obvious' that the agent tried to perform an action. It means that we abstract away from any intentional and/or volitional elements of attempts which are completely internal to the agents. It also means that we can only state that an agent attempts to perform an action if it at least has the capability to perform the action. This is again
justified by the organizational perspective on actions. If e.g. a university would let a
sociology lecturer try to teach a class on nuclear physics and he failed we would not say that this was a
serious attempt, because it is obvious that this person does not have the capability to teach this class
(even though at an individual level the sociology lecturer might genuinely have intended to teach the class,
formed intentions, etc.). This aspect is even more obvious in agent organizations. An agent should only
attempt to see to it that a situation holds if it knows which actions it has to perform and it has these
capabilities (designed in its program, action or plan base). So, although we have used variants of dynamic
logic to characterize attempts of actions before \cite{wieringa:96} and recently a new logic of attempt is
described in \cite{lorini:08}, these are very useful to describe attempts from an individual's point of view,
but not from an organizational perspective.

Abstract logics of agency start by defining the modal operator $E$ for direct, successful action, and
introduce the other operators, for capability, ability, attempt or indirect action, based on the definition
of $E$. This results in axioms such as $E_a\varphi \to C_a\varphi$, informally meaning that if $a$ sees to it
that $\varphi$ then $a$ is (cap)able of $\varphi$. From a realistic perspective, such a definition is pretty
uninteresting. Obviously, if you indeed do something, you must have had the ability to do it, but the
interesting issue is to, given one's capabilities, determine under which circumstances one can reach a
certain state of affairs. That is, to determine in which situations it can be said that $C_a\varphi$ leads to $E_a\varphi$. For instance, if agent $a$ is capable of achieving $\varphi$ and also responsible and no other
agent is interfering then agent $a$ will actually achieve $\varphi$.

Our approach is thus to start by giving the definition of agent capability and use this definition to
progressively introduce the definitions for ability, attempt and activity. We furthermore provide a semantic
definition of the modal operators instead of the usual axiomatic definition. The idea is that a semantic
definition helps to precisely identify all the wanted characteristics, while a sound axiomatic system can be
given on top of these definitions. Of course, it is also possible to start with an axiomatic system and
construct a (canonical) semantic framework. However, in this case the semantics will add little to our
understanding of the fundamental properties of the modal operators. Finally, we also agree with
\cite{governatori:03} that the assumption taken in \cite{santos-jones-carmo:97} that indirect action always
implies an impossibility for direct action is rather strong, and will not use it.

Intuitively, the ability of an agent to realize a state of affairs $\varphi$ in a world $w$, depends not only
on the capabilities of the agent but also on the status of that world. Therefore, we define the ability of
$a$, $G_a\varphi$ to represent the case in which the agent has not only the potential capability to establish
$\varphi$ but is currently in a state in which it has influence over some of the possible transitions that
lead to a state where $\varphi$ holds. Thus the agent also has an actual possibility to use its capability.
The attempt by agent $a$ to realize $\varphi$ is represented by $H_a\varphi$. An agent attempts to realize
$\varphi$ if $\varphi$ holds after all states that can be reached by a transition that is influenced by $a$.
In our definition of attempt, an attempt only fails in case another agent interferes and tries achieve
something which prevents $\varphi$ to be achieved. So, we do not consider that the environment might just
prevent $\varphi$ to be achieved with some probability. We also assume an agent only attempts to achieve
things it is capable of achieving (which does not necessarily mean the agent \emph{knows} it is capable of
achieving them).  In the special case in which all next possible states from a given state are influenced by
an agent $a$, we say that $a$ is \emph{in-control} in $w$, represented by $IC_a$. Finally, the \emph{stit}
operator, $E_a\varphi$ ('agent $a$ sees to it that $\varphi$) represents the result of successful action
(that is, $\varphi$ holds in all worlds following the current one). This notion of agent activity is based on
that introduced by P\"orn \cite{porn:74} to represent the externally \emph{`observable'} consequences of an
action instead of the action itself, and as such abstracts from internal motivations of the agents.
\emph{Stit} can be seen as an
abstract representation of the family of all possible actions that result in $\varphi$.\\

Formally, the agent operators are defined as follows, where all definitions are introduced as an extension to
CTL*. Given a set $A$ of agents, we start by defining the language $\mathcal{L}_O$ for LAO as an extension of
$\mathcal{L}$ as follows:
\begin{definition}\label{def:languageLO}
Given a set $A$ of agents, $\mathcal{L}_O$ is such that
\begin{enumerate}
    \item $\varphi \in \mathcal{L} \Rightarrow \varphi \in \mathcal{L}_O$
    \item $a \in A, \varphi \in \mathcal{L}_O \Rightarrow C_a\varphi, G_a\varphi, H_a\varphi, E_a\varphi, IC_a \in \mathcal{L}_O$
\end{enumerate}
\end{definition}
Semantics to all of these operators are build over the semantic notion of capabilities. As discussed above,
the intuition is that an agent possesses capabilities that make action possible. That is, in order to talk
about agent activity, or, that agent $a$ possesses the ability to make proposition $\varphi$ hold in some
next state in a path from the current world, we need to establish the control of the agent over the (truth)
value of $\varphi$. For each agent $a$, we partition the set of atomic facts $\Psi$ in any world $w$ of M in
two classes: the set of atomic facts that agent $a$ can control, $c(a,w)$, and the set of atomic
facts that $a$ cannot control, $\Psi \backslash c(a,w)$. 
Given a set $c(a,w)$ of propositional capability of $a$, we  define $\Sigma_a \subseteq \mathcal{L}_O$
inductively as follows:\\
\indent
\begin{tabular}{l}
    $\forall p \in c(a,w)$, $\pi(p) \in \Sigma_a$,\\
    $\forall p \in c(a,w)$, $\neg \pi(p) \in \Sigma_a$,\\
    $\forall \psi_1, \psi_2 \in \Sigma_a$, $\psi_1\wedge\psi_2 \in \Sigma_a$ iff $\not\models \psi_1 \to \neg\psi_2$ and $\not\models \psi_2 \to \neg\psi_1$\\
\end{tabular}\\
Agents can control
the state of $\varphi$ in a world, represented by $C_a\varphi$ iff the agent
controls the set of atomic facts that yield $\varphi$ true. Furthermore, LAO is
defined such that no agent can control the obvious (tautologies), and if an
agent controls a fact $p$ it also controls its negation $\neg p$.  
\begin{definition}[Agent Capability]\label{def:agentcapability}
Given an agent $a \in A$, agent capability, $C_a\varphi$ is defined as: $w \models C_a\varphi$ iff $\exists
w': w \neq w', w' \models\neg\varphi$ and $\exists \psi \in \Sigma_a : \models \psi \rightarrow \varphi$
\end{definition}
\begin{definition}[Agent Ability, Attempt and Activity]\label{def:agentcapability2}
Given an agent $a \in A$, agent ability, $G_a\varphi$, agent attempt, $H_a\varphi$, agent control, $IC_a$, and agent activity, $E_a\varphi$, are defined as:\\
\indent
\begin{tabular}{lccl}
   $G_a$: & $w \models G_a\varphi$ & iff & $w \models C_a\varphi$ and $\exists (w,w') \in T_{aw}: w'\models \varphi$\\
  $H_a$: & $w \models H_a\varphi$ & iff & $w \models G_a\varphi$ and $\forall (w,w') \in T_{aw}: w'\models \varphi$\\
  $IC_a$: & $w \models IC_a$ & iff & $\forall w': (w,w') \in Rt \Rightarrow (w,w') \in T_{aw}$\\
  $E_a$: & $w \models E_a\varphi$ & iff & $w \models H_a\varphi \wedge IC_a$
\end{tabular}
\end{definition}
Note that these definitions include the notion that action takes time by representing the result of action in a following world.
\begin{figure}[ht]
\centering\epsfig{file=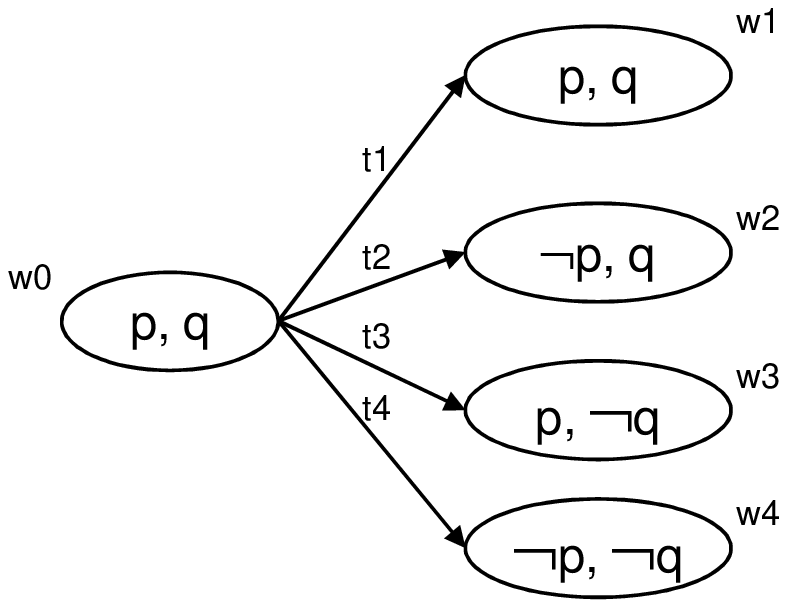,width=.3\textwidth,angle=270}
\caption{Example for capability, ability and attempt.}
\label{fig:excontrol}
\end{figure}

As an example, consider model $M = (\{p,q\}, \{a\}, \{w_0,...,w_4\}, \{t_1,...,t_4\}, T,\pi, c(a,
w_i)=\{p\})$, where $i = 0, ..., 4$,
as depicted in figure \ref{fig:excontrol}. We furthermore define that $(w_0, w_1)\in T_{aw_0}$. In this
model, it holds that:\\
\indent
\begin{tabular}{l}
    $w_0 \models G_ap$ (because $C_ap$ and $w_1 \models p$)\\
    $w_0 \models H_ap$ if $T_{aw} = \{(w_0,w_1), (w_0,w_3)\}$\\
    $w_0 \not\models C_a(p\wedge q)$\\
    $w_0 \not \models G_a(p\wedge q)$ (because $\neg C_a(p\wedge q)$) \\
\end{tabular}  \\

\subsubsection{Group Activity}
Agents are, by definition, limited in their capabilities, that is, the set of facts in the world that they
can control. This implies that certain states of affairs can only be reached if two or more agents cooperate
to bring that state into existence. We define control and action of a group of agents based on the combined
atomic capabilities of the agent in the group, using the normal set theoretic notions. I.e.
$c(Z,w)=\bigcup_{a\in Z} c(a,w)$ and $T_{Zw}=\bigcup_{a\in Z} T_{aw}$.
In the same way as for single agents, we define composed capability $\Sigma_Z$ of a group $Z$ as follows:\\
\indent
\begin{tabular}{l}
    $\forall p \in C_Z$, $p \in \Sigma_Z$\\
    $\forall p \in C_Z$, $\neg p \in \Sigma_Z$\\
    $\forall \psi_1, \psi_2 \in \Sigma_S$, $\psi_1\wedge\psi_2 \in \Sigma_Z$ iff $\not\models \psi_1 \to \neg\psi_2$ and $\not\models \psi_2 \to
    \neg\psi_1$\\
\end{tabular}\\

Table \ref{table:operators} gives an overview of the semantics of the $C, G, H$ and $E$ operators for groups
of agents. We refer the reader to \cite{dignum:omas:chapter:09} for a more extensive formal specification.
\begin{table}[h]\centering \caption{Overview of group operators and their semantics}\label{table:operators}
\begin{tabular}{|p{2cm}|c|p{9.0cm}|}
 \hline
  & \textbf{Operator} & \textbf{Definition} \\
 \hline
 \textbf{Capability} & $C_Z\varphi$ & $w \models C_Z\varphi$ iff $\exists w': w' \models\neg\varphi$ and $\exists \psi \in \Sigma_Z$ such that $w \models \psi \rightarrow \varphi$\\
 \hline
 \textbf{Joint Capability} & $JC_Z\varphi$ & $w \models JC_Z\varphi$ iff $w \models C_Z\varphi$ and $\forall Z' \subset Z, w \models \neg C_{Z'}\varphi$\\
 \hline
 \textbf{Ability} & $G_Z\varphi$ & $w \models G_Z\varphi$ iff $w \models C_Z\varphi$ and $\exists (w,w') \in T_{Zw}: w'\models \varphi$\\
            &              & where $T_{Zw} = \bigcup\limits_{a\in Z} T_{aw}$\\
 \hline
 \textbf{Attempt} & $H_Z\varphi$ & $w \models H_Z\varphi$ iff $w \models G_Z\varphi$ and $\forall (w,w') \in T_{Zw}: w'\models \varphi$\\
 \hline
 \textbf{In-control} & $IC_Z$ & $w \models IC_Z$ iff $\forall w': (w,w')\in R \Rightarrow \exists a\in Z: a\in T(w,w')$\\
 \hline
 \textbf{Stit}  & $E_Z\varphi$ & $w \models E_Z\varphi$ iff $w \models H_Z\varphi \wedge IC_Z$\\
 \hline
\end{tabular}
\end{table}

\section{Organization Structure in LAO}\label{sect:lao-org}
The idea behind organizations is that there are global objectives that can only be achieved through combined
agent action. One of the main reasons for creating organizations is to provide the means for coordination that enables the achievement of organizational objectives in an efficient manner. From the definition of
group capabilities, it easily follows that groups of agents have at least as many capabilities as each of their members
and thus groups are more ``powerful''. In order to achieve its objectives, it is necessary that besides employing
the relevant agents, the organization assures that their interactions and responsibilities enable an
\emph{efficient} realization of its objectives.

According to this view of organization, even if the agents in the organization have group control over all
organizational objectives, they still need to coordinate their activities in order to efficiently achieve
those objectives. Furthermore, in most cases, the objectives of the organization are only known to a few of
the agents in the organization, who may have no control over those objectives. It is therefore necessary to
organize the agents in a way that enables objectives to be passed to those agents that can effectively
realize them. We need therefore to extend our groups to include coordinating and task allocation concepts.

Formally this is achieved by adding a set of \emph{organizational structures} $Os=\{O_1,...,O_n\}$ to the
model that can be used to evaluate constructs that describe them. Each organizational structure $O_i$ is a
tuple
\begin{equation}\label{eq:lao-org}
  O_i = (As_i,R_i,rea_i,\leq_i,D_i,Obj_i,K_i)
\end{equation}
describing the elements of an organization, as follows:
\begin{itemize}
    \item $As_i$ : $W \rightarrow 2^A$ indicates in each state the set of agents belonging to organization $O_i$
    \item $R_i$ : $W \rightarrow 2^R$ indicates in each state the set of roles belonging to organization $O_i$ where $R$ stands for the (finite) set of all possible roles
    \item $rea_i$ : $W \rightarrow (As_i \times R_i)$ indicates in each state which agents play which role(s)
    \item $\leq_i$ : $W \rightarrow (R_i \times R_i)$ defines the dependency relation between roles in an organization
    \item $D_i$ : $W \rightarrow 2^{\Psi}$ indicates the desired state of the organization in each world
    \item $Obj_i$ : $R_i \times W \rightarrow 2^{(\Psi \backslash \top)}$ indicates which role in the organization needs to realize a certain objective of the organization
    \item $K_i=(K^+_i,K^-_i)$ with:\\
    $K^+_i$ : $W \rightarrow 2^{\Psi}$ indicates the positive ``knowledge'' of the organization\\
    $K^-_i$ : $W \rightarrow 2^{\Psi}$ indicates the negative ``knowledge'' of the organization
\end{itemize}

\noindent The function $As_i$ defines the agents that participate in the organization at each point in time.
It determines the workforce and thus also the capabilities available to the organization.

\noindent The function $R_i$ defines the roles of the organization. Although we assume that the set of roles
is more stable than the set of agents participating in the organization, we keep open the possibility for
reorganization by having the set of roles depend on the state.

\noindent The function $rea_i$ relates the agents to the roles they play at each moment in time. Note that, at any time,
agents can play more than one role and roles can be played by more than one agent.

\noindent The relation $\leq_i$ indicates the \emph{structural dependency} between the set of roles of
organization $O_i$. This relation can be seen as a kind of power relation between the agents playing the
roles in an organization\footnote{LAO abstracts from the different types of power that exist in real organizations and furthermore does not elaborate further the consequences of power in communication between agents. More on this aspect can be found in \cite{dignum:coinecai:06}.}. Distribution of tasks is done following these dependency structures. $\leq_i$ is a
poset satisfying the following properties:
\begin{enumerate}
    \item $\forall w, \forall r \in R_i(w) : \leq_i(r,r,w)$ (reflexivity)
    \item $\forall w, \forall p,r,q \in R_i(w)$ : if $\leq_i(p,r,w)$ and $\leq_i(r,q,w)$ then
        $\leq_i(p,q,w)$ (transitivity)
\end{enumerate}
The reflexivity property makes sure that each role has power over itself and the transitivity property makes
sure that an agent can distribute tasks to agents that are somewhere below it in the dependency
structure\footnote{In reality, this transitivity is not always automatic, i.e. the chain of command must be followed to delegate tasks downwards. By this formalization, is meant the final distribution of tasks resulting from this chain of commands.}. Because we also want to be able to talk about distribution of tasks by an agent to a group of
other agents we generalize the dependency relation to groups of roles, as follows:\\
\indent
\begin{tabular}{l}
    $\leq_i(r,S,w)$ iff $\forall q \in S \subseteq R_i(w): \leq_i(r,q,w)$ \\
    $\leq_i(S,S',w)$ iff $\forall q \in S', \exists r \in S: \leq_i(r,q,w)$  \\
\end{tabular}  \\

Intuitively, the ordering relation in the set of roles determines the
interaction possibilities between agents playing the roles. Organizational structures influence
the way that agents in the organization can interact.  The relation $r \leq_i
q$ indicates that an agent playing $r$ is able to interact with an agent playing $q$ in order to request or
demand some result. $\leq_i(r,S,w)$ means that agents playing role $r$ can demand a result from agents playing any of the roles in $S$. E.g. the CEO of a company can order both the sales manager as well as the production manager to achieve some goal (such as increase number of products sold). Finally, $\leq_i(S,S',w)$ means that for an agent $i$ fulfilling a role from $S'$ one can find a role in $S$ such that an agent $j$ fulfilling that role can demand a result from $i$. E.g. for every person working in a construction company (carpenter, plumber, etc.) there is a manager that supervises that person. This construction makes it possible to say that the board of managers is managing the workers, while not every manager can direct all workers. In this paper, we will not further detail the
types of interactions between agents (delegation, request, bid, ...) but assume
that the relationship will achieve some result, through a more or less complex
interaction process. More on this issue can be found in
\cite{dignum:coinecai:06}.

The set of domain facts given by $D_i$ indicates the state that the organization $O_i$ tries to achieve. If
$M,w \models D_i(w)$ the organization has achieved its objective.\\
The relation $Obj_i$ indicates which role is responsible (through the agents playing it) to initiate actions
to achieve a certain objective of the organization (given by a set of domain facts, which cannot include
$\top$). We say that an organization is \emph{well-defined} iff $D_i(w) \subseteq \bigcup_{r\in R_i}
Obj_i(r,w)$, i.e. for all desired objectives of the organization there is a role in charge of
reaching that objective.\\
The last relation $K_i$ denotes the ``knowledge" of the organization. In this paper, we assume this
knowledge to be information that is available to all agents in the organization. It can be seen as a central
database. We mainly introduce this element because we want to be able to describe situations where agents
might have certain capabilities, but this is not known in the organization and thus a task is not delegated
to the right agent. In \cite{dignum:coinecai:06} we discuss how the structural dependencies in an
organization can influence the flow of information in that organization. As before, we take here the
incremental approach, using $K_i$ as a `placeholder' for more complex treatment of organizational
information. 
We also explicitly do not model this operator as a traditional modal epistemic relation which would require a
thorough discussion of all possible connections between this modal operator and the action related operators.
This is also left for future research. We do, however, distinguish $K^+_i$ and $K^-_i$ in order to be able to
distinguish between the fact that something is known to be true, known to be false or not known. We also
require that $K^+_i(w) \subset w$ which means that all facts that are known are also true and $K^-_i(w)\cap
w=\emptyset$ which means that things that are known to be false do not hold in the world.

%

Given the above organizational concepts, we can now extend the
language $\mathcal{L_O}$. First we introduce some predicates that describe various organizational relations.
\begin{definition}\label{def:languageLO-org}
Given an organization $O_i = (A_i, R_i, rea_i, \leq_i, D_i, Obj_i, K_i)$, the organization language $\mathcal{L_O}$ introduced in Definition \ref{def:languageLO}, is extended to include the following predicates, where
$a\in A_i$,$r,q\in R_i$ and $\\varphi \in \mathcal{L_O}$:\\
\indent
\begin{tabular}{l}
  \textit{member}$(a,O_i) \in \mathcal{L_O}$, \\
  \textit{role}$(r,O_i) \in \mathcal{L_O}$, \\
  \textit{play}$(a,r,O_i) \in \mathcal{L_O}$,  \\
  \textit{dep}$(O_i,r,q) \in \mathcal{L_O}$, \\
  \textit{know}$(O_i, \varphi) \in \mathcal{L_O}$, \\
  \textit{incharge}$(O_i,r,\varphi) \in \mathcal{L_O}$ with $\varphi$ not containing negations,   \\
  \textit{desire}$(O_i,\varphi) \in \mathcal{L_O}$ with $\varphi$ not containing negations \\
\end{tabular}\\
with the following semantics:\\
\begin{tabular}{l}
  $M_O, w \models member(a,O_i)$ iff $a \in As_i(w)$,\\
  $M_O, w \models role(r,O_i)$ iff $r \in R_i(w)$, \\
  $M_O, w \models play(a,r,O_i)$ iff $rea_i(w,a,r)$,\\
  $M_O, w \models dep(O_i,r,q)$ iff $r\leq_i(w) q$,\\
  $M_O, w \models know(O_i,p)$ iff $\pi(p)\in K^+_i(w)$ for $p\in \Phi$,\\
  $M_O, w \models know(O_i,\neg p)$ iff $\pi(p)\in K^-_i(w)$ for $p\in \Phi$,\\
  $M_O, w \models know(O_i,\varphi\wedge\psi)$ iff $M_O, w\models know(O_i,\varphi)$ and\\
  \indent $M_O, w \models know(O_i,\psi)$,\\
  $M_O, w \models incharge(O_i,r, p)$ iff $\pi(p)\in Obj_i(r,w)$ for $p\in \Phi$,\\
  $M_O, w \models desire(O_i,p)$ iff $\pi(p)\in D_i(w)$ for $p\in \Phi$,\\
  $M_O, w \models incharge(O_i,r,\varphi\wedge\psi)$ iff $M_O, w\models incharge(O_i,r,\phi)$ and\\
  \indent $M_O, w \models incharge(O_i,r,\psi)$\\
  $M_O, w \models desire(O_i,\varphi\wedge\psi)$ iff $M_O, w\models desire(O_i,\varphi)$ and\\
  \indent $M_O, w \models desire(O_i,\psi)$\\
\end{tabular}\\
\end{definition}
We do not allow negations within the scope of $incharge$ and $desire$ in order to keep the semantics of these
concepts relatively simple. \footnote{If we would allow negations we could e.g. not use simple sets of desires to define the semantics, because this does not provide a means to express a difference between $desire(O_i,\neg p)$ and $\neg desire(O_i, p)$.} In most practical cases it also appears sufficient to model all that we want as most desires and objectives are formulated as positive facts that are to be established.
The same facts can be defined for groups of roles. In particular, given an organization
structure and the definition of the dependency relation between roles we can also define a similar relation
of dependency on a group of roles as follows:
\begin{definition}[Group Dependency]\label{def:chain}
For $O_i = (As_i, R_i, rea_i, \leq_i, D_i, Obj_i, K_i)$, and group of roles $U \subseteq
R_i$, \textit{dep}($O_i$,r,U) is defined as\\
\indent
\begin{tabular}{l}
    $dep(O_i,r,Z) \equiv \forall q \in Z: dep(O_i,r,q)$.\\
\end{tabular}
\end{definition}

Besides the above extensions we also want to extend the language with organizational actions, i.e, the
actions performed by agents playing a role in the organization.
\begin{definition}
The organization language $\mathcal{L_O}$ in Definition \ref{def:languageLO-org}, is extended, for
organization $O_i = (As_i, R_i, rea_i, \leq_i, D_i, Obj_i, K_i)$, to include the following predicates.
$\forall a\in A$, $\forall r\in R$, and $\forall \varphi \in \mathcal{L_O}$: \\
\indent
\begin{tabular}{l}
  $C_{ar}\varphi \in \mathcal{L_O}$,\\
  $G_{ar}\varphi \in \mathcal{L_O}$,\\
  $H_{ar}\varphi \in \mathcal{L_O}$,\\
  $IC_{ar} \in \mathcal{L_O}$,\\
  $E_{ar}\varphi \in \mathcal{L_O}$\\
\end{tabular}\\
\end{definition}
Where e.g. $C_{ar}\varphi$ stands for the fact that $\varphi$ is part of the capabilities of the role $r$
enacted by agent $a$. In order to be able to give semantics to these types of actions we will change the
semantic model with respect to actions slightly. Mainly we couple the agent labels in the model to roles. A
restriction is that agents can only be coupled to roles they are playing. We assume that everything an agent
does is done in its capacity of enactor of a certain role. So, agents do not act without having a role attached to it.
We can still talk about an agent performing an action if the role is not important. This is made possible through the second clause in definition \ref{def:transitioninfluence2}. We could also facilitate this by having a special role (something like "agent") which has no necessary capabilities and can be played by any agent in any state and denotes that the agent is not acting as playing an organizational role. We do not pursue this further in this paper, because we are mainly interested in agents functioning completely within the organizational setting.\\
Within $M_O$ we change the relation $T$ such that it relates tuples of agents and roles to transitions and
the capabilities are now also defined for roles and tuples of agents playing roles:
\begin{itemize}
\item $c: A \times W \rightarrow 2^{\Psi}$ associates agents with facts in $\Psi$, i.e. $c(a)$
    indicates the capabilities of $a$
\item $cn: R \times W \rightarrow 2^{\Psi}$ associates roles with facts in $\Psi$, indicating the necessary capabilites that an agent should have to play role $r$.
\item $cr: A \times R \times W \rightarrow 2^{\Psi}$ is a function that associates tuples (agent,role) in
    a world w with the facts in $\Psi$
\item $T: Rt \rightarrow 2^{A\times R}$ is the set of (agent,role) labels on elements of $Rt$,
\end{itemize}
The function $cn$ restricts the relation $rea_i$ such that $rea_i(w,a,r)$ can only hold if $cn(r, w)\subseteq
c(a, w)$. I.e. an agent $a$ can only play a role $r$ if it has the necessary capabilities to play that role.
Furthermore, $cr(a,r,w)$ is only defined iff $rea_i(w,a,r)$.
We assume that agents can get extra capabilities when playing roles and thus that $c(a,w)\subseteq
cr(a,r,w)$. The set $\Sigma_{ar}$ is defined inductively over $cr$ in the same way as $\Sigma_a$ was defined
over $c$.\\
Roles only appear in the transition influence relation when played by an agent: $(a,r)$ can only be an element of $T((w,w'))$ iff $rea_i(w,a,r)$ holds.

We also redefine the notion of transition influence as follows:
\begin{definition}[Transition influence]\label{def:transitioninfluence2}
Given a model $M_O = (\Psi, A, W, Rt, T, \pi, c)$, a world $w \in W$ and an agent $a \in A$, role $r \in R$
such that $rea_i(a,r,w)$, the transition influence of ($a,r$) in $w$, $T_{arw}$ and $T_{aw}$ are defined by:\\
\indent \begin{tabular}{l}
    $T_{arw} = \{rt | rt \in Rt: \exists w': rt = (w,w')$ and $(a,r) \in T(rt)\}$\\
    $T_{aw} = \{rt | rt \in Rt: \exists w': rt = (w,w')$ and $\exists r \in R$: $(a,r) \in T(rt)\}$\\
\end{tabular}
\end{definition}
Given the definitions above we can keep all the definitions on agent capability, ability, attempt and
activity. But we now have additional definitions for agents playing roles.
\begin{definition}[Role enacting Agent Capability, Ability, Attempt, Control and Activity]\label{def:agentcapability3}
Given agent $a \in A$,$r \in R$, role enacting agent capability, $C_{ar}\varphi$, role enacting agent
ability, $G_{ar}\varphi$, role enacting agent attempt, $H_{ar}\varphi$, role enacting agent control,
$IC_{ar}$, and role enacting agent activity, $E_{ar}\varphi$, are defined as:\\
\indent
\begin{tabular}{lccl}
  $C_{ar}$: & $w \models C_{ar}\varphi$ & iff & $\exists w'\not w: w' \models\neg\varphi$ and $\exists \psi \in \Sigma_{ar}$ such that $\models \psi \rightarrow \varphi$\\
  $G_{ar}$: & $w \models G_{ar}\varphi$ & iff & $w \models C_{ar}\varphi$ and $\exists (w,w') \in T_{arw}: w'\models \varphi$\\
  $H_{ar}$: & $w \models H_{ar}\varphi$ & iff & $w \models G_{ar}\varphi$ and $\forall (w,w') \in T_{arw}: w'\models \varphi$\\
  $IC_{ar}$: & $w \models IC_{ar}$ & iff & $\forall w': (w,w') \in Rt \Rightarrow (w,w') \in T_{arw}$\\
  $E_{ar}$: & $w \models E_{ar}\varphi$ & iff & $w \models H_{ar}\varphi \wedge IC_{ar}$
\end{tabular}
\end{definition}
A few comments are due concerning the above definitions. For the purpose of this paper, we see agents as
actors of roles in a organization, that is, the agent's objectives are those of the roles they enact. We assume that, by acting according to their capabilities, agents work towards organizational objectives. We assume that agents have their own internal motivations to enact a role \cite{dastani:03}, but that motivation is not discussed further in this paper. In fact, the notion of agent in LAO is that of role-enacting agent as described in \cite{virginia:phd:04}. Furthermore,
organizational knowledge is treated, for the moment, as a label marking facts that are explicitly known by
the organization. That is, we do not provide epistemic capabilities in order to reason about knowledge,
rather than asserting the fact that something is known. Future extensions to LAO will improve these initial assumptions.

Like with actions of agents we can extend the above definitions to groups of role enacting agents:
\begin{definition}[Groups of Role enacting Agents Capability, Ability, Attempt, Control and Activity]\label{def:agentcapability4}
Given a set of agents $V \subseteq A$ and a set of roles $U \subseteq R$, role enacting agent capability,
$C_{ar}\varphi$, role enacting agent ability, $G_{ar}\varphi$, role enacting agent attempt, $H_{ar}\varphi$,
role enacting agent control, $IC_{ar}$, and role enacting agent activity, $E_{ar}\varphi$, are defined as:\\
\indent
\begin{tabular}{lcp{1cm}p{7cm}}
  $C_{VU}$: & $w \models C_{VU}\varphi$ & iff & $\exists w' \models\neg\varphi$ and $\exists a\in V$ $\exists r \in U$ $\exists \psi \in \Sigma_{ar}$ such that $w \models \psi \rightarrow \varphi$\\
  $G_{VU}$: & $w \models G_{VU}\varphi$ & iff & $w \models C_{VU}\varphi$ and $\exists a\in V$ $\exists r \in U$ $\exists (w,w') \in T_{arw}: w'\models \varphi$\\
  $H_{VU}$: & $w \models H_{VU}\varphi$ & iff & $w \models G_{VU}\varphi$ and $\forall a\in V$ $\forall r \in U$ if $rea_i(w,a,r)$  then $\forall (w,w') \in T_{arw}: w'\models \varphi$\\
  $IC_{VU}$: & $w \models IC_{VU}$ & iff & $\forall w': (w,w') \in Rt \Rightarrow (\exists a\in V \exists r\in U: rea_i(w,a,r)$ and $(w,w') \in T_{arw}$)\\
  $E_{VU}$: & $w \models E_{VU}\varphi$ & iff & $w \models H_{VU}\varphi \wedge IC_{VU}$
\end{tabular}
\end{definition}

\subsection{Organization Capability}
Based on the definitions given in the previous sections, we now define organization capability (or scope of
control). In fact, an organization is only as good as its agents. An organization is said to be capable of
achieving $\varphi$ if there is a subset of agents in that organization that has the capability to achieve
$\varphi$. Of course, whether an organization actually is able to achieve $\varphi$ also depends on whether
the task of achieving this objective arrives at the agents that are capable of achieving it. We will treat that
aspect later on. Formally the capability of an organization is defined as follows:
\begin{definition}[Organization Capability]\label{def:orgcapability}
Given a model $M_O$, a world $w \in W$ and organization $O_i$  organizational capability $C_{O_i}$ is defined
as:\\
\indent
\begin{tabular}{l}
    $M_O, w \models C_O\varphi$ iff $\exists Z \subseteq As_i(w): M_O, w \models C_Z\varphi$\\
\end{tabular}
\end{definition}

\subsection{Responsibility and Initiative}
To further refine the concept of organization, we need to be able to describe and reason about the
responsibilities that are linked to the roles within the organization. Responsibilities within an
organization enable agents to make decisions about what each member of the organization is expected to do, according to the organization specification, and to anticipate the tasks of others \cite{grossi:07}. Informally, by responsibility we mean that an agent
or group has to make sure that a certain state of affairs is achieved, either by realizing it itself or by
delegating that result to someone else. Extensive work on responsibility has been done, amongst others, by
Jones and Sergot \cite{jonessergot:96} and by Castelfranchi \cite{castelfranchi:95}, which provides a richer
description of the concepts of responsibility and power. In line with our aim to provide a comprehensive but
minimal formal framework to reason about organization and change, in the remainder of this section we focus on the issue of initiative taking as part of responsibility. I.e. we concentrate on the case that, if the organization has defined the responsibilities of to include some activity, agents playing that role have to initiate some action towards that activity. They cannot wait until they are asked by others to perform some task. On the other hand, the organization assumes that the agents playing the role will eventually do something about the activity that the role is responsible for\footnote{In this paper, we don't develop further on the issue of eventual action.}.\\
We introduce the operator $incharge$ to represent the case that a certain role is in
charge (has the initiative) of accomplishing a certain state.
In order to describe this notion of initiative from the role perspective, we introduce
a new operator, $I_r$, such that $I_r\varphi$ means that $r$ has the initiative to achieve $\varphi$, which means that an agent playing $r$ should perform some action to achieve $\varphi$. The initiative operator is also
defined for a group $Z$ of roles, $I_Z$, in a similar way.
\begin{definition}
Formally, we extend the language $\mathcal{L}_O$ as follows:\\
\indent
\begin{tabular}{l}
    $r \in R, \varphi \in \mathcal{L}_O \Rightarrow I_r\varphi \in \mathcal{L}_O$ \\
       $Z \subseteq R, \varphi \in \mathcal{L}_O \Rightarrow I_Z\varphi \in \mathcal{L}_O$ \\
\end{tabular}\\
\end{definition}
Before we give a semantic definition of Initiative we first show how initiative is related to
\textit{incharge($O_i,r,\varphi$)} previously defined facts:
\begin{equation}\label{eq:initiative}
  \models \textit{incharge}(O_i,r,\varphi) \rightarrow I_r\varphi
\end{equation}
The validity expresses that if a role defined by the organization to be responsible to achieve $\varphi$ it will
automatically take the initiative to act such that $\varphi$ will be achieved. Because roles do not act
themselves, this means that an agent playing the role $r$ will act in order to achieve $\varphi$. In
a richer model this would of course also involve the decision process of the agent(s), including forming the decision of delegating or performing $\varphi$ itself. In that case, the
definition below would be extended by a weaker version that states that if the role is in charge of $\varphi$
than an agent enacting that role is \emph{obliged} to take initiative to achieve $\varphi$. This normative
aspect of the responsibility is quite important, but also very complex in combination with delegations in
organizations. See e.g. \cite{lima:10,limaetal:10} for more details on this matter. Again, we leave the
combination of that theory with the organizational framework for future work.

Note that when an agent will take initiative to achieve a state of affairs this does not guarantee successful
achievement of that state of affairs. In fact, initiative can be defined informallu as: role $r$ has initiative to achieve $\varphi$ iff an agent $a$ playing $r$  will eventually attempt to achieve $\varphi$ or attempt to put another role in charge of $\varphi$. Formally initiative is defined in terms of attempt, as follows:
\begin{definition}[Initiative]\label{def:responsibility}
Given model $M_O$ and organization $O_i$, $O_i = (As_i,R_i,rea_i,\leq_i, D_i, Obj_i, K_i)$, initiative $I_r\varphi$, resp. $I_Z\varphi$, for role $r \in R_i(w)$, resp. group $Z \subseteq R_i(w)$, is defined as:\\
\indent
\begin{tabular}{lp{0.6cm}p{9cm}}
  $w \models I_r\varphi$ & iff & $w \models \exists a: play(a,r,O_i) \wedge \lozenge( H_{ar}\varphi \vee H_{ar} incharge(O_i,q,\varphi))$, for some $q \in R_i(w)$\\ 
  $w \models I_Z\varphi$ & iff & $\exists U \subseteq As_i(w) \forall a\in U\exists r\in Z:  w \models play(a,r,O_i) \wedge \lozenge( H_{UZ}\varphi \vee H_{UZ} incharge(O_i,Z',\varphi))$, for some $Z' \subseteq R_i(w)$\\ 
\end{tabular} \\
\end{definition}
Note that the above definition requires that if a role has the initiative to achieve a certain state
$\varphi$ there also is at least one agent playing that role (which can then try to achieve $\varphi$). From this agent it is required that eventually it attempts to achieve $\varphi$ or that it attempts to put another role in charge of achieving $\varphi$. Whether the agent can attempt to "delegate" an initiative to another role depends on the relations there exists between the roles. In the next section we will define the notion of delegation that indicates when such an attempt is certainly possible.
Different properties can be defined for the initiative operator, which identify different types of
organizations. For example, a \emph{well-defined organization} is one where there is someone in charge for
each of the organizational objectives.
\begin{definition}[Well-Defined Organization]\label{def:welldef-organization}
Given an organization $O_i$  in a model $M_O$, $O_i = (As_i,R_i,rea_i, \leq_i, D_i, Obj_i, K_i)$, we say that
$O_i$ is a \emph{well-defined organization} if it satisfies the following requirement:
\begin{equation}
    M_O,w \models \textit{desire}(O_i,\varphi) \rightarrow \exists r: (\textit{role}(r,O_i) \wedge I_r\varphi)
\end{equation}
\end{definition}
We call an organization \emph{successful} if the organization also has the capabilities to achieve each
objective. Formally,
\begin{definition}[Successful Organization]\label{def:success-organization}
Given an organization $O_i$  in a model $M_O$, $O_i = (As_i,R_i,rea_i, \leq_i, D_i, Obj_i, K_i)$, we say that
$O_i$ is a \emph{successful organization} if it satisfies the following requirement:
\begin{equation}
    M_O,w \models \textit{desire}(O_i,\varphi) \rightarrow C_{O_i}\varphi \wedge \exists r: (\textit{role}(r,O_i) \wedge I_r\varphi)
\end{equation}
\end{definition}
Note that, due to the definition of $I_r$, this definition of success is dependent on the specific agents that play the roles in the organization.

\subsection{Delegation}
Delegation of tasks is defined as the capability to put an agent, or group, in charge for that task (through
the roles they play). In an organization, the power of delegation is associated with structural dependencies.
That is, by nature of their dependencies, some agents are capable of delegating their tasks to other agents,
through the respective roles they play in the organization.  Formally,
\begin{definition}[Power of delegation]\label{def:delegation}
Given an organization $O_i$  in a model $M_O$, $O_i = (As_i,R_i,rea_i, \leq_i, D_i, Obj_i, K_i)$, the power
of delegation for $\varphi$ between two roles $r, q \in R_i(w)$ is defined as the following constraint in the
model:\\
\indent
\begin{tabular}{l}
    if $M_O,w\models \textit{dep}(O_i,r,q)\wedge \textit{incharge}(O_i,r,\varphi) \wedge play(a,r,O_i) $\\
    then $M_O,w\models C_{ar}\textit{incharge}(O_i,q,\varphi)$\\ 
\end{tabular}
\end{definition}
%
Given the notions of initiative and structural dependency introduced above, we can define a \emph{good
organization} as follows:
\begin{definition}[Good Organization]\label{def:goodorganization}
Given an organization $O_i$  in a model $M_O$, $O_i = (As_i,R_i,rea_i, \leq_i, D_i, Obj_i, K_i)$, we say that
$O_i$ is a \emph{good organization} if it satisfies the following requirement:\\
\indent
\begin{tabular}{l}
  if $w \models (C_{O_i}\varphi \wedge I_Z\varphi)$ then \\
  $(\exists U\subseteq R_i(w)$  and $V=\{a\|rea_i(w,a,r), r\in U\}$ and $ w \models \textit{dep}(O_i,Z,U) \wedge C_{VU}\varphi)$\\
\end{tabular}\\

\noindent where $Z, U \subseteq R_i(w)$ represent a group of roles in $O_i$ and $V$ is defined as the set of
agents playing one of the roles of the set $U$.
\end{definition}
That is, a \emph{good organization} is an organization such that if the organization has the capability to
achieve $\varphi$ and there is a group of roles in the organization responsible for realizing it, then the
roles being in charge have a chain of delegation to roles that are played by agents in $As_i$ that are
actually capable of achieving it. Note that it is possible that $Z=U$. \emph{Good organizations} satisfy the
following property:
\begin{equation}
    I_r\varphi \rightarrow \lozenge H_{O_i}\varphi
\end{equation}
which informally says that if there is a role in charge of a given state of affairs, then eventually the
state will be attempted (of course, the success of such attempt is dependent on possible external
interferences). This follows immediately from the previous definition. We define an \emph{efficient}
organization
as:\\
\begin{definition}[Efficient Organization]\label{def:org-efficient}
Given an organization $O_i$  in a model $M_O$, $O_i = (As_i,R_i,rea_i, \leq_i, D_i, Obj_i, K_i)$, we say that
$O_i$ is a \emph{efficient organization} if it satisfies the following requirement:\\
\indent
\begin{tabular}{llp{10cm}}
  $M_O,w$ & $ \models $ & $(I_r\varphi \wedge (\forall a \in A_i: play(a,r,O_i) \rightarrow \neg C_{ar}\varphi) \wedge (\exists q\in R_i:\textit{dep}(O_i,r,q)) \wedge (\exists b\in A_i: \textit{play}(b,q,O_i)) \wedge \textit{know}(O_i, C_{bq}\varphi)) \rightarrow 
  E_{ar} \textit{incharge}(O_i,q,\varphi)$\\
\end{tabular}\\
\end{definition}
This states that if a role $r$ is in charge to achieve $\varphi$ but none of the agents playing role $r$ is
capable, but it is known that an agent $b$ playing role $q$ does have the capability to achieve $\varphi$ and
also $q$ is a subordinate role of $r$ then some agent $a$ playing role $r$ will delegate the responsibility
to role $q$. Thus $a$ hands responsibility of tasks to those agents of which it is known that they can
achieve them (if it cannot achieve them itself).

\subsection{Supervision}
Related to the notion of a good organization is the idea that agents should supervise each other's work. I.e. if role $r$ is in charge that agent $b$ attempts to achieve a certain objective then role $r$ becomes responsible for that objective again if $b$ fails in his attempt. For instance, when a project leader delegates the task of implementing a module of the system to a certain person and that person fails to implement the module (maybe because he becomes ill) then the project leader should take back the task and give it to someone else (or do it himself). Note that we talk about the project leader being a role in the organization. In principle it does not matter which person plays that role of project leader. The person always has to act when the programmer fails. We can define this property as follows:
\begin{definition}[Supervising Duty]\label{def:supervising}
Given an organization $O_i$  in a model $M_O$, $O_i = (As_i,R_i,rea_i,\leq_i, D_i, Obj_i, K_i)$, and group of
roles $Z\subseteq R_i(w)$, and a group of agents $V\subseteq As_i(w)$ playing the roles $U\subseteq R_i(w)$,
the supervising duty of roles $Z$ with respect to the group agents $V$ to realize $\varphi$ is defined as:\\
\begin{tabular}{c}
    $SD_{(Z,V)}\varphi \equiv (I_Z H_{VU}\varphi\wedge \lozenge(H_{VU}\varphi\wedge X\neg\varphi))\rightarrow I_Z\varphi)$.
\end{tabular}
\end{definition}
The definition states that if $V$ attempts to realize $\varphi$ with a certain degree of uncertainty while
under the responsibility of $Z$, then roles $Z$ becomes directly in charge of achieving $\varphi$ every time
$V$ fails to realize $\varphi$.

\section{On the axiomatization of LAO}\label{sect:analysis}
In the previous sections we have introduced both syntax and semantics of LAO. In order to provide a logical
characterization of LAO, in this section we discuss an axiomatization of LAO and give a number of theorems.
The soundness of the axioms follows directly from the definitions and thus we abstain from giving formal
proofs, and only indicate proof sketches where necessary. We also do not pretend that the given axiom system
is complete, which would require a separate paper due to the intricate relations between the modal
operators introduced.\\
Moreover, we give most axioms on the operators applied to single agents. These axioms can also be formulated
for groups of agents. We leave these out as they would just duplicate all axioms without adding any new
intuition.

We give the axioms in sets that are related to the same intuition. The first important intuition is that we
do not want agents to have capabilities to achieve tautologies. I.e. achievements should be related to state
of affairs whose truth value can be influenced and thus can be either true or false. This leads to the
following axioms.

\paragraph{\textbf{Axioms}}
\begin{description}
   \item[\texttt{(A1)}] $\neg C_a \top$
   \item[\texttt{(A2)}] $\neg C_ar \top$
   \item[\texttt{(A3)}] $\neg incharge(O_i,r,\top)$
\end{description}
The soundness of follows directly from the definitions. The other achievement operators are defined over the
capability operator and thus they ``inherit" the property directly. This can be seen after we presented the
axioms relating the operators a bit further below.

Given these axioms the next question obviously is whether the achievement modalities are ``normal" modalities
in the sense that the K axiom holds. Having the K axiom allows to at least combine achievements or attempts
in order to reason about the combinations. Fortunately we still have that for most of them:

\paragraph{\textbf{Axioms}}
\begin{description}
   \item[\texttt{(A4)}] $C_a \varphi \wedge C_a \psi \rightarrow C_a(\varphi \wedge\psi)$
   \item[\texttt{(A4r)}] $C_{ar} \varphi \wedge C_{ar} \psi \rightarrow C_{ar}(\varphi \wedge\psi)$
   \item[\texttt{(A5)}] $H_a \varphi \wedge H_a \psi \rightarrow H_a(\varphi \wedge \psi)$
   \item[\texttt{(A5r)}] $H_{ar} \varphi \wedge H_{ar} \psi \rightarrow H_{ar}(\varphi \wedge \psi)$
   \item[\texttt{(A6)}] $I_r \varphi \wedge I_r \psi \rightarrow I_r(\varphi \wedge \psi)$
   \item[\texttt{(A7)}] $incharge(O_i,r,\varphi) \wedge incharge(O_i,r,\psi) \rightarrow
       incharge(O_i,r,\varphi \wedge \psi)$
   \item[\texttt{(A8)}] $desire(O_i,\varphi) \wedge desire(O_i,\psi) \rightarrow desire(O_i,\varphi
       \wedge \psi)$
\end{description}
We also have K axiom property for $E_a$ and $E_{ar}$, however because these are defined in terms of $H_a$ and
$IC_a$ and $H_{ar}$ and $IC_{ar}$, these properties are given as theorems:

\paragraph{\textbf{Theorems}}
\begin{description}
   \item[\texttt{(T1)}] $\vdash E_a \varphi \wedge E_a \psi \rightarrow E_a(\varphi \wedge \psi)$
   \item[\texttt{(T2)}] $\vdash E_{ar} \varphi \wedge E_{ar} \psi \rightarrow E_{ar}(\varphi \wedge
       \psi)$
\end{description}
Only for abilities this works the other way (due to the existential quantifier in the definition of its
semantics):

\paragraph{\textbf{Axioms}}
\begin{description}
   \item[\texttt{(A9)}] $G_a(\varphi \wedge \psi)\rightarrow G_a \varphi \wedge G_a \psi$
   \item[\texttt{(A9r)}] $G_{ar}(\varphi \wedge \psi)\rightarrow G_{ar} \varphi \wedge G_{ar} \psi$
\end{description}
There are some ways in which achievements of agents and achievements of role enacting agents can be related.
First note that the capabilities of an agent should include the capabilities necessary to play a role, but
that an agent enacting a role can have more capabilities than an agent not enacting a role. Thus:

\paragraph{\textbf{Axioms}}
\begin{description}
   \item[\texttt{(A10)}] $\forall r,a,O_i: (play(a,r,O_i)\wedge C_r\varphi)\rightarrow C_a\varphi$
   \item[\texttt{(A11)}] $\forall r,a,O_i: (play(a,r,O_i)\wedge C_a\varphi)\rightarrow C_{ar}\varphi$
\end{description}
The other achievement operators of agents and role enacting agents are related as follows:

\paragraph{\textbf{Axioms}}
\begin{description}
   \item[\texttt{(A12)}] $C_a\varphi\wedge G_{ar}\varphi \rightarrow G_a\varphi$
   \item[\texttt{(A13)}] $\forall r,a,O_i: (play(a,r,O_i)\wedge H_a\varphi) \rightarrow H_{ar}\varphi$
   \item[\texttt{(A14)}] $IC_{ar} \rightarrow IC_a$
\end{description}
Because role enacting agents can have more capabilities axiom (A12) only holds because we require
$C_a\varphi$. Note that we also do not have $G_a\varphi \rightarrow G_{ar}\varphi$, because $a$ might have
influence on some transition enacting another role but no influence on any transition while enacting role
$r$.\\
Axiom (A13) follows from the fact that $T_{arw}\subseteq T_{aw}$ and thus if something holds over all
transitions in $T_{aw}$ it also holds for all subsets of transitions and in particular for $T_{arw}$ for any
$r$ that is enacted by $a$. Axiom (14) follows from the same fact.\\
From the above axiom we can easily see that the following theorem is true:

\paragraph{\textbf{Theorems}}
\begin{description}
   \item[\texttt{(T3)}] $\vdash (IC_{ar}\varphi\wedge H_{ar}\varphi) \rightarrow H_a\varphi$
\end{description}
So, if a role $r$ enacting agent $a$ is in control and attempting to achieve $\varphi$ then $a$ is also
attempting to achieve $\varphi$ (because there are no transitions in which $a$ has influence and in which he
is not enacting $r$, due to the fact that $IC_{ar}$).\\
 We can now also prove a property similar to the above
axioms to relate $E_a$ and $E_{ar}$:

\paragraph{\textbf{Theorems}}
\begin{description}
   \item[\texttt{(T4)}] $\vdash C_a\varphi\wedge E_{ar}\varphi \rightarrow E_a\varphi$
\end{description}
The proof follows easily from axioms (A12), (A14) and the theorem above. We do not have the T axiom for any
of these modalities. The only axiom that resembles the T axiom is stating that if an agent (enacting a role)
sees to it that $\varphi$ this will be true in the next state:

\paragraph{\textbf{Axioms}}
\begin{description}
   \item[\texttt{(A15)}] $E_a\varphi \rightarrow X \varphi$
   \item[\texttt{(A15r)}] $E_{ar}\varphi \rightarrow X \varphi$
\end{description}
From this axiom it is also right away clear that we don't have the usual theorem that follows from the T
axiom:
\begin{equation*}
E_a E_a\varphi \not\rightarrow E_a \varphi
\end{equation*}
The antecedent translates to: $a$ sees to it that (in the next state) it will see to it that $\varphi$. So,
then $\varphi$ will be realized only in two time steps from the current state. The consequent requires
$\varphi$ to be achieved in the next time step already. So, the fact that in our framework the achievement
takes time prevents this theorem to be true and unnesting of achievement operators is not possible. We do
have the T axiom (and the K axiom) for the {\emph know} relation:

\paragraph{\textbf{Axioms}}
\begin{description}
   \item[\texttt{(A16)}] $know(O_i,\varphi) \rightarrow \varphi$
   \item[\texttt{(A17)}] $know(O_i,\varphi) \wedge know(O_i,\psi) \rightarrow know(O_i,\varphi
       \wedge\psi)$
\end{description}
We do not have the usual axioms 4 and 5, because we do not allow nesting of the {\emph know} operator.\\
Finally, we do have the D axiom (and the K axiom) for organizational desires:

\paragraph{\textbf{Axioms}}
\begin{description}
   \item[\texttt{(A18)}] $\neg desire(O_i,\perp)$
   \item[\texttt{(A19)}] $desire(O_i,\varphi) \wedge desire(O_i,\psi) \rightarrow desire(O_i,\varphi
       \wedge\psi)$
\end{description}
There are a number of axioms that relate the different achievement operators for agents and also similar ones
relating role enacting agent achievement operators:

\paragraph{\textbf{Axioms}}
\begin{description}
   \item[\texttt{(A20)}] $G_a\varphi \rightarrow C_a\varphi$
   \item[\texttt{(A20r)}] $G_{ar}\varphi \rightarrow C_{ar}\varphi$
   \item[\texttt{(A21)}] $H_a\varphi \rightarrow C_a\varphi$
   \item[\texttt{(A21r)}] $H_{ar}\varphi \rightarrow C_{ar}\varphi$
   \item[\texttt{(A22)}] $H_a\varphi \rightarrow G_a\varphi$
   \item[\texttt{(A22r)}] $H_{ar}\varphi \rightarrow G_{ar}\varphi$
   \item[\texttt{(A23)}] $E_{ar}\varphi \rightarrow I_r\varphi$
   \item[\texttt{(A24)}] $H_{ar}\varphi \rightarrow I_r\varphi$
   \item[\texttt{(A25)}] $incharge(O_i,r,\varphi) \rightarrow I_r\varphi$
\end{description}
Axioms (A20-A22) are standard and follow directly from the definitions of the semantics of the operators.
Axioms (A23) and (A24) are maybe less obvious. They state that role enacting agents only attempt or see to it
that a state $\varphi$ becomes true if the role they are playing has the initiative to attempt or achieve
that state. This means that role enacting agents only perform actions that are supporting the objectives of
the roles they enact. This does not mean that agents cannot perform any other actions, but, that in their role
enacting capacity they are limited to perform actions for which the role has the initiative. This property
gives some handles to make some statements about an organization based on its desires and the objectives of
the roles abstracting away from which precise agents fulfill the roles. Given these axioms we can easily
prove the following theorems:

\paragraph{\textbf{Theorems}}
\begin{description}
   \item[\texttt{(T5)}] $\vdash \neg I_r \top$
   \item[\texttt{(T6)}] $\vdash \neg G_a \top$
   \item[\texttt{(T7)}] $\vdash \neg G_{ar} \top$
   \item[\texttt{(T8)}] $\vdash \neg H_a \top$
   \item[\texttt{(T9)}] $\vdash \neg H_{ar} \top$
   \item[\texttt{(T10)}] $\vdash \neg E_a \top$
   \item[\texttt{(T11)}] $\vdash \neg E_{ar} \top$
   \item[\texttt{(T12)}] $\vdash E_a\varphi \rightarrow C_a\varphi$
   \item[\texttt{(T13)}] $\vdash E_{ar}\varphi \rightarrow C_{ar}\varphi$
   \item[\texttt{(T14)}] $\vdash E_a\varphi \rightarrow H_a\varphi$
   \item[\texttt{(T15)}] $\vdash E_{ar}\varphi \rightarrow H_{ar}\varphi$
\end{description}
The first seven follow from the relation between $G,H$ and $E$ to $C$ and the fact that we have $\neg
C_a\top$ (A1). The last four follow directly from the definition of $E$.

The following theorems relate potential achievements of different agents at the same time:

\paragraph{\textbf{Theorems}}
\begin{description}
       \item[\texttt{(T16)}] $\vdash E_a\varphi \rightarrow \forall b:\neg G_b\neg\varphi$
       \item[\texttt{(T17)}] $\vdash E_ar\varphi \rightarrow \forall b:\neg G_b\neg\varphi$
       \item[\texttt{(T18)}] $\vdash E_ar\varphi \rightarrow \forall q:\forall b:\neg G_bq\neg\varphi$
\end{description}
Again the proof is direct from the definition of the $E$ operator. Given that agent $a$ (possibly enacting a
role $r$) sees to it that $\varphi$ is achieved means that $a$ is in control. If $a$ is in control and
attempts to achieve $\varphi$ then in all states following the current one holds $\varphi$. As a consequence
there is not one following state where $\neg\varphi$ can hold and thus it is not possible for any other agent
to have the ability to achieve $\neg\varphi$ in the current state. Given the above theorems the following
theorems are obvious from the definitions of the operators as well:

\paragraph{\textbf{Theorems}}
\begin{description}
   \item[\texttt{(T19)}] $\vdash E_a\varphi \rightarrow \neg E_b\neg\varphi$
   \item[\texttt{(T20)}] $\vdash E_ar\varphi \rightarrow \forall b:\neg E_b\neg\varphi$
   \item[\texttt{(T21)}] $\vdash E_ar\varphi \rightarrow \forall q:\forall b:\neg E_bq\neg\varphi$
   \item[\texttt{(T22)}] $\vdash E_a\varphi \rightarrow \neg H_b\neg\varphi$
   \item[\texttt{(T23)}] $\vdash E_ar\varphi \rightarrow \forall b:\neg H_b\neg\varphi$
   \item[\texttt{(T24)}] $\vdash E_ar\varphi \rightarrow \forall q:\forall b:\neg H_bq\neg\varphi$
\end{description}
Only when an agent sees to it that $\varphi$ becomes true it is not possible for another agent to try to
achieve its negation. Note that, the following theorems that could be expected as in other theories do not hold here:\\
\indent
\begin{tabular}{l}
   $\not\vdash H_a\varphi \rightarrow \neg H_b\neg\varphi$\\
   $\not\vdash H_{ar}\varphi \rightarrow \forall b:\neg H_b\neg\varphi$\\
   $\not\vdash H_{ar}\varphi \rightarrow \forall q:\forall b:\neg H_{bq}\neg\varphi$\\
\end{tabular} \\
which would indicate the impossibility of parallel \emph{attempts} to achieve $\varphi$ and its negation by
different agents (possibly enacting roles)! Finally, we add the following two axioms relating agents to
organizations:

\paragraph{\textbf{Axioms}}
\begin{description}
   \item[\texttt{(A26)}] $member(a,O_i) \wedge C_a\varphi \rightarrow C_{O_i}\varphi$
   \item[\texttt{(A27)}] $incharge(O_i,r,\varphi) \wedge dep(O_i,r,q) \rightarrow \exists a:
       play(a,r,O_i) \wedge C_a incharge(O_i,q,\varphi)$
\end{description}
We do not have the necessitation rules for the operators but we do have the following rules of inference:

\paragraph{\textbf{Rules of inference}}
\begin{description}
   \item[\texttt{(R1)}] If $\models \varphi \leftrightarrow \psi$ then $\models C_a\varphi
       \leftrightarrow C_a\psi$
   \item[\texttt{(R2)}] If $\models \varphi \leftrightarrow \psi$ then $\models C_{ar}\varphi
       \leftrightarrow C_{ar}\psi$
   \item[\texttt{(R3)}] If $\models \varphi \leftrightarrow \psi$ then $\models G_a\varphi
       \leftrightarrow G_a\psi$
   \item[\texttt{(R4)}] If $\models \varphi \leftrightarrow \psi$ then $\models G_{ar}\varphi
       \leftrightarrow G_{ar}\psi$
   \item[\texttt{(R5)}] If $\models \varphi \leftrightarrow \psi$ then $\models H_a\varphi
       \leftrightarrow H_a\psi$
   \item[\texttt{(R6)}] If $\models \varphi \leftrightarrow \psi$ then $\models H_{ar}\varphi
       \leftrightarrow H_{ar}\psi$
   \item[\texttt{(R7)}] If $\models \varphi \leftrightarrow \psi$ then $\models E_a\varphi
       \leftrightarrow E_a\psi$
   \item[\texttt{(R8)}] If $\models \varphi \leftrightarrow \psi$ then $\models E_{ar}\varphi
       \leftrightarrow E_{ar}\psi$
   \item[\texttt{(R9)}] If $\models \varphi \leftrightarrow \psi$ then $\models I_r\varphi
       \leftrightarrow I_r\psi$
\end{description}
With these rules we conclude the sketch of an axiomatization for the LAO framework. As said before, we do not
attempt to give a complete study of the LAO framework as a logic, which we leave for a separate paper. Rather
we will now continue to show how the framework can function as a basis to specify organizational change
(through operators that change the organization from one state to another) and how this helps to specify and
analyze different use cases in a uniform way.

\section{Application of LAO}\label{sec:application}
So far, we have introduced a formal model to represent organizations. Our claim
is that such models enable the analysis and modelling of organizations in a
formal and realistic way, and are as such more qualified for organization
representation than (classic) logics. In this section, we will further discuss
this claim by using the model to analyze abstract organization types and by
applying it to the case study introduced in section \ref{section:scenario}.
LAO has also been applied to formally describe other existing agent systems, such as Adaptive Architectures for Command and Control (described in \cite{dignum:omas:chapter:09}),  RoboSoccer (described in \cite{dignum:jancl:10}),
and crisis management (described in \cite{penserini:aose:09}). Furthermore, LAO is used as the formal basis
for the OperettA tool for design and analysis of organization models~\cite{dignum:operetta:10}. Together,
this provides a basis to demonstrate the power of LAO to model and analyze different types of organizations and study
organizational changes.

\subsection{Types of Organizations}
The work of Mintzberg on the analysis of organizational structures
\cite{mintzberg:93} is widely used within Organization Theory. According to
Mintzberg, environmental variety is determined by both environmental complexity
and the pace of change. Inspired by and extending the work of Mintzberg,
researchers in Organizational Theory have proposed a growing number of
classifications for organization structures, e.g. simple, bureaucracy, matrix,
virtual enterprize, network, boundary-less organizations, conglomeration,
alliance, etc. just to name a few forms. However, definitions and naming of
organizational forms are often unclear and the classification of a specific
organization into one of the proposed classes is not trivial, often resulting
in hybrid forms.

Based on the structure of control, Burns and Stalker distinguish two main types of organization structures:
hierarchies and networks \cite{burns:61}. Using the formal definition of organization presented in the
previous sections, we are able to specify the structural characteristics of these two types of organizations.

In a hierarchy, organizational objectives are known to the managers who delegate them to their subordinates who
have the capabilities and resources to realize them. Formally,
\begin{definition} \textbf{(Hierarchy)}\\
A structured organization $O_i = (As_i,R_i,rea_i,\leq_i,D_i,Obj_i,K_i)$ is said to be an \emph{hierarchy} iff
$\forall w \in W$:\\
\indent
\begin{tabular}{l}
    $\exists M \subset R_i, M \neq \oslash: \forall m \in M, \exists \varphi \in D_i: \varphi \in Obj_i(m,w) \wedge$ \\
    \indent $\forall \varphi \in D_i, \exists m \in M: \varphi \in Obj_i(m,w)$, and \\
    $\forall r \in R_i, r \not\in M, \exists m \in M: \leq_i(m,r,w)$\\
\end{tabular}\\
\end{definition}
That is, there is a group of managers $M$ that together is responsible for all organizational objectives and have
a chain of command to all other agents in the organization. In a simple, flat, hierarchy,
the manager group furthermore meets the following requirements:\\
\indent
$M = \{m\}$ and $\forall r \in R_i: \leq_i(m,r,w)$ and $\not\exists s \in R_i, s \neq m: \leq_i(s,r,w)$.\\

In the same way, we formally define a network organization as:
\begin{definition} \textbf{(Network)}\\
A structured organization $O_i = (As_i,R_i,rea_i,\leq_i,D_i,Obj_i,K_i)$ is said to be a \emph{network} iff
$\forall w \in W$:\\
\indent
\begin{tabular}{l}
    $\forall r \in R_i: Obj_i(r,w) \cap D_i \neq \oslash$ and $\forall \varphi \in D_i: \exists r \in R_i: \varphi \in Obj_i(r,w)$  \\
    $\forall r \in R_i, \exists s \in R_i: \leq_i(r,s,w)$\\
\end{tabular}\\
\end{definition}
That is, every agent in the organization is responsible for some of the organizational objectives, and have a
delegation relationship to some other agent in the organization. A fully connected network
also meets the following requirement:\\
\indent $\forall r, s \in R_i: \leq_i(r,s,w)$.\\
A team is also a special case of network that meets the symmetry requirement:\\
\indent $\leq_i(r,s,w) \rightarrow \leq_i(s,r,w)$.

\subsection{Case study}
We have applied LAO to formally describe the gas market organization introduced
in section \ref{section:scenario}. The situation before liberalization
is described by:\\

$O^0_{gas} = (As^0_{gas},R^0_{gas},rea^0_{gas},\leq^0_{gas}, D^0_{gas}, Obj^0_{gas}, K^0_{gas})$, where
\begin{itemize}
  \item $As^0_{gas} = \{m, t, s, l\}$ are the agents, with capabilities\\
   $C_t\emph{buy-gas}, C_s\emph{transport-gas}, C_l\emph{local-flow}$
  \item $R^O_{gas} = \{\textit{monopolist}, \textit{trader}, \textit{shipper}, \textit{local-transport}\}$
      are the roles
  \item $rea^0_{gas} = \{(m,\textit{monopolist}), (t,\textit{trader}), (s,\textit{shipper}),
      (l,\textit{local-transport})\}$ are the initial role enacting agent relations
  \item $\leq^0_{gas} = \{\textit{monopolist}\leq_{gas} \textit{trader},\textit{monopolist}\leq_{gas}
      \textit{shipper},\textit{monopolist}\leq_{gas} \textit{local-transport}\}$ specifies the control of
      \textit{monopolist} roles over the other partners in the supply chain
  \item $D^0_{gas} = \{\emph{provide-gas}\}$
  \item $Obj^0_{gas}(\textit{monopolist}) = \{\emph{provide-gas}\}$
  \item $Obj^0_{gas}(\textit{trader}) = \{\emph{buy-gas}\}$
  \item $Obj^0_{gas}(\textit{shipper}) = \{\emph{transport-gas}\}$
  \item $Obj^0_{gas}(\textit{local-transport}) = \{\emph{local-transport}\}$
  \item $K^0_{gas} =  \{know(O_{gas}, \emph{provide-gas} \leftarrow (\emph{buy-gas} \wedge
      \emph{transport-gas} \wedge \emph{local-flow})\}$ are the known means to achieve the objective of
      providing gas.
\end{itemize}

It is straightforward to see that this organization is in a state in which it can realize its objective of
providing gas. In fact, according to definition \ref{def:welldef-organization}, this is a \emph{well-defined
organization}, since there is a role that has the objective of realizing the desires of the organization. Formally:\\

\begin{tabular}{lr}
    $\textit{desire}(O_{gas},\textit{provide-gas})$ & (from $O_{gas}$)\\
    $\textit{incharge}(\textit{monopolist}, \textit{provide-gas})$ & (from $O_{gas}$)\\
    $\textit{incharge}(\textit{monopolist}, \textit{provide-gas}) \rightarrow I_{\textit{monopolist}}\textit{provide-gas}$ & (from (4))\\
    $\therefore I_{\textit{monopolist}}\textit{provide-gas}$ & \\
    $\therefore \textit{desire}(O_{gas},\textit{provide-gas}) \rightarrow I_{\textit{monopolist}}\textit{provide-gas}$ & \\
    $\therefore \textit{desire}(O_{gas},\textit{provide-gas}) \rightarrow \exists r : \textit{role}(r, O_{gas}) \wedge I_{r}\textit{provide-gas}$ & \\
\end{tabular}\\

\noindent In the same way can be checked that $O_{gas}$ is a \emph{successful organization} (definition
\ref{def:success-organization}) since all capabilities needed to achieve the desire of $O_{gas} ($\textit{provide-gas}) are available and there is a role that has the initiative to achieve the desire:

\begin{tabular}{lr}
    $C_t\emph{buy-gas} \wedge C_s\emph{transport-gas} \wedge C_l\emph{local-flow}$ & (from $O_{gas}$)\\
    $\therefore C_{O_{gas}}\textit{provide-gas}$ & (def. \ref{def:orgcapability} and def. of $C_Z$)\\
    $I_{\textit{monopolist}}\textit{provide-gas}$ & (from above)\\
    $\therefore \textit{desire}(O_{gas}, \textit{provide-gas}) \rightarrow$ & \\
    $C_{O_{gas}}\textit{provide-gas} \wedge \exists r : \textit{role}(r, O_{gas}) \wedge I_r\textit{provide-gas}$  \\
\end{tabular}\\

The proof that $O_{gas}$ is also a \emph{good organization} (definition \ref{def:goodorganization}) follows
directly from the fact that the monopolist has delegation power over the agents that have the capabilities to
realize the organization's objectives.

However, as in all monopolies, the process is fully determined and controlled by the monopolist and dependent
on him.
Following the political decision on the liberalization of the gas market in the
Netherlands, the role of the monopolist disappears. This means that, unless
other changes are made, the organization as described above is no longer in
state of achieving its objectives. This is due to the fact that delegation
power relations no longer exist and furthermore no agent is responsible for the
organizational objective $D_O$. Typically, such situation will demand extra
capabilities from the parties involved in order to coordinate their activities
without control by the monopolist. A possible solution is to evolve into a
network organization, which provides the flexibility needed to deal with this
situation. create the possibility for the other partners to directly contract
each others and for multiple parties to enter the market. The resulting
situation can be described as follows:

$O^{0'}_{gas} = (As^{0'}_{gas},R^{0'}_{gas},rea^{0'}_{gas},\leq^{0'}_{gas}, D^{0'}_{gas}, Obj^{0'}_{gas}, K^{0'}_{gas})$, where:
\begin{itemize}
  \item $As^{0'}_{gas} = \{t, s, l\}$ are the agents, with capabilities as above
  \item $R^{0'}_{gas} = \{\textit{trader}, \textit{shipper}, \textit{local-transport}\}$ are the roles
  \item $rea^{0'}_{gas} = \{(t,\textit{trader}), (s,\textit{shipper}), (l,\textit{local-transport})\}$ are the initial role enacting agent relations
  \item $\leq^{0'}_{gas} = \{\textit{trader}\leq^{0'}_{gas} \textit{shipper},\textit{trader}\leq^{0'}_{gas} \textit{local-transport},\textit{shipper}\leq^{0'}_{gas} \textit{trader},\\
      \textit{shipper}\leq^{0'}_{gas} \textit{local-transport},\textit{local-transport}\leq^{0'}_{gas}\textit{trader},\\
      \textit{local-transport}\leq^{0'}_{gas}\textit{shipper}\}$ specify the role dependencies
  \item $D^{0'}_{gas} = \{\emph{provide-gas}\}$
  \item $Obj^{0'}_{gas}(\textit{trader}) = \{\emph{buy-gas},\emph{provide-gas}\}$
  \item $Obj^{0'}_{gas}(\textit{shipper}) = \{\emph{transport-gas},\emph{provide-gas}\}$
  \item $Obj^{0'}_{gas}(\textit{local-transport}) = \{\emph{local-transport},\emph{provide-gas}\}$
  \item $K^{0'}_{gas} =  \{know(O_{gas}, \emph{provide-gas} \leftarrow (\emph{buy-gas} \wedge
      \emph{transport-gas} \wedge \emph{local-flow})\}$ are the known means to achieve the objective of providing gas.
\end{itemize}
That is, the monopolist is removed, the roles of trader, shipper and local transport are extended to take the
desire  of the organization, \emph{provide-gas}, as their objective, and dependencies between all roles are
created such that a peer network relation is achieved. These network relations enable the parties to
negotiate service contracts between themselves. Each of the role enacting agents will be in a state of engaging the others such that they can achieve \emph{provide-gas}, as for example in the following sequence of states:\\
\noindent
\begin{tabular}{|lp{7.6cm}p{0.3cm}p{3cm}|}
  \hline
  $s_{1_a}$: & $\emph{incharge}(O_{gas},\emph{trader},\emph{provide-gas})$ & & (from $O_{gas}$)\\
  $s_{1_b}$: & $I_{\emph{trader}}(\emph{buy-gas} \wedge \emph{transport-gas}\wedge \emph{local-flow})$ & & (from $O_{gas}$)\\
  $s_{1_c}$: & $C_{(t,\emph{trader})}(\emph{incharge}(O_{gas},\emph{shipper}, \emph{transport-gas})\wedge C_{(t,\emph{trader})}(\emph{incharge}(O_{gas},\emph{local-tranport}, \emph{local-flow})$& & (from $O_{gas}$ and \textit{incharge} def.)\\
  $s_2$: & $E_{(t,\emph{trader})}\textit{incharge}(O_{gas},\emph{shipper}, \emph{transport-gas})$ & & (for some $s_2>s_1$)\\
  $s_3$: & $E_{(t,\emph{trader})}\textit{incharge}(O_{gas},\emph{local-tranport}, \emph{local-flow})$ & & (for some $s_3>s_1$)\\
  $s_{4_a}$: & $I_{\emph{trader}}\emph{buy-gas} \wedge C_{(t,\emph{trader})}\emph{buy-gas}$ & & (properties of E and \textit{incharge})\\
  $s_{4_b}$: & $I_{\emph{shipper}}\emph{transport-gas} \wedge C_{(s,\emph{shipper})}\emph{transport-gas}$ & & (properties of E and \textit{incharge})\\
  $s_{4_c}$: & $I_{\emph{local-tranport}}\emph{local-flow} \wedge C_{(l,\emph{local-tranport})}\emph{local-flow}$ & & (properties of E and \textit{incharge})\\
  $s_5$: & $H_{(t,\emph{trader})}\emph{buy-gas}$  & & (for some $s_5>s_4$)\\
  $s_6$: & $H_{(s,\emph{shipper})}\emph{transport-gas}$  & & (for some $s_6>s_4$)\\
  $s_7$: & $H_{(l,\emph{local-tranport})}\emph{local-flow}$  & & (for some $s_7>s_4$)\\
  $s_8$: & $(\emph{buy-gas} \wedge \emph{transport-gas}\wedge \emph{local-flow})$  & & \\
  & $\therefore \emph{provide-gas}$ & &\\
  \hline
\end{tabular}    \\
  \\

Note that $O'_{gas}$ does not have knowledge of the capabilities of the agents, and therefore does not comply
with the notion of efficient organization as in definition \ref{def:org-efficient}. In the above example of
an operation, we do assume that knowledge of the capabilities of role enacting agents is available, and as such the trader passes the initiative for parts of the operation to the `correct' agents right away. This is merely to illustrate the operation. Without this knowledge the organization would probably end up trying to put roles in charge of states
of affairs for which the agents playing them would not have the capabilities. Only by an exhaustive trial and error the right distribution of initiatives would be found. We furthermore left out of the example the fact
that initiative leads to attempt, which may or not result in a successful activity. In the example above all
attempts succeed.

This example demonstrates that LAO is capable of representing different types
of organizations and can be used to analyze their characteristics. However, it
does not as yet provide the means to represent the actual change from one to
the other organization structure. In order to be able to specify and analyze
the process of moving from organization $O$ to organization $O'$, the LAO model
needs to be extended with mechanisms to describe organizational change. The dynamics of LAO are described in \cite{dignum:jancl:10}.

\section{Related Work}\label{sect:related}
Several approaches have  been presented to investigate the complexity of
reasoning and analysis of multi-agent systems. Even though taking diverse perspectives, all approaches
are concerned with some of the requirements presented in section \ref{sect:intro} above, and can be basically divided
into two categories: formal methods and engineering frameworks.

Formal methods for MAS have a logical basis, typically based on dynamic, temporal and/or deontic logics \cite{hoek-wooldridge:05,santos-jones-carmo:97,governatori:03}. However, often their treatment of organizational concepts is in the sense that only specific issues are considered.
The approach described in \cite{hoek-wooldridge:05} is based on ATL. In ATL logics, an operator is defined that expresses that a coalition of agents controls a formula. I.e. $\ll C \gg \phi$ means that the coalition C can achieve $\phi$ no matter what the agents outside C do. It is a kind of ensured achievement. However, in order to be able to ensure this in a consistent way assume a complete division of agent capabilities and total control over the domain. This means that all basic propositions can be controlled by exactly one agent. Although this provides a nice logical system it is not very realistic as usually a proposition can be controlled by more than one agent. In LAO, it is possible to express the possible interference of the agents starting with basic propositions.
The work presented by Santos et.al. in
\cite{santos-jones-carmo:97} also uses attempts to achieve situations and therefore is closer to our approach in this respect than ATL. However, that work lacks temporal issues. Therefore it is not possible to reason about the difference that an action makes on the world. E.g. in LAO we can state $\neg \phi \wedge E_a\phi$ stating that $\phi$ is false, but agent $a$ will make it true. In a \emph{stit} logic without temporal aspects it holds that $E_a\phi \rightarrow \phi$ and thus the formula above would be inconsistent.
Popova and Sharpanskykh \cite{popova:09} describe a ontological framework for modeling and analysis of organizations based on TTL. The framework includes concepts and relations partitioned into four dedicated views: process-oriented, performance-oriented, organization-oriented and agent-oriented view. The framework can be used for representing different types of organizations ranging from mechanistic to organic. A similar approach is that by Horling and Lesser \cite{horling:08}

Besides the formal, logical approaches towards organizational description and analysis there are also engineering frameworks. They provide sound representation languages that
include many realistic organizational concepts, but have often a limited formal
semantic basis, which makes analysis and comparison difficult
\cite{hubner:ooop:06,mccallum:06,broek:06}. LAO is able to provide a basis for formal semantics to these engineering approaches. We already showed that it is possible to formalize some organizations described in quite different approaches.

Some approach that also combines theory and practice is that about electronic institutions. See \cite{esteva:01} for a description of a formal approach to describe electronic institutions. This seems quite close to the work on organizations, because institutions are also used to regulate the interactions between autonomous agents. However, there are some important differences between organizations and institutions that lead to very different requirements on their formal specification. The first difference is that the agents that interact through an institution do not have to be aware of any objective of the institution. They interact purely to try to satisfy their own objectives as good as possible. Therefore the objectives of the institution do not have to be explicit and the actions of the agents do not have to be linked to them. It also means that the delegation of parts of objectives does not appear (explicitly) in institutions. Thus we do not need concepts such as power relations, delegation and responsibility.\\
The second important difference is that institutions assume total control of the interactions within the institution. Therefore the actions of the agents do not interfere with each other. In organizations we assume that agents have more autonomy, because they are meant to take care of (at least part of) the objectives of the organization. Local knowledge and perspectives, different capabilities and power relations lead to possibly inconsistent attempts of agents to achieve the objectives of the organization. This perspective of possible interference of actions of agents lead to a large part of the logical framework dealing with the possible interference of actions and thus action attempts instead of only successful actions. This is not necessary in institutions where the institution allows only those actions that lead to a successful outcome.\\
So, we can conclude that, although there is certainly an overlap between institutions and organizations, they have different foci and requirements leading to quite different requirements for their formal specification and thus the resulting formal frameworks are by right quite different as well.

Computational organizational science tries to combine
the organization theory and engineering framework perspectives. It looks at
groups, organizations and societies and aims to understand, predict and manage
system level change \cite{carley:02:pnas}. Several tools for the analysis and
modelling of organizations have been proposed. Computational models, in
particular those based on representation techniques and empirical simulation,
have been widely used for several decades to analyze and solve organizational
level problems. More recently, mathematical tools, including those of decision
and game theory, probability and logic are becoming available to handle
multiple agency approaches to organizations.
In practice, organizational level solutions are provided by mathematical and
computational models based on probabilistic and decision theoretic approaches.
A large body of work in this area is that of computational simulation,
specifically agent-based social simulation (ABSS) \cite{davidsson:02}.
Computational simulations are based on formal models, in the sense that they
provide a precise theoretical formulation of relationships between variables.
Simulations  provide a powerful way to analyze and construct realistic models
of organizational systems and make possible to study problems that are not
easily addressed by other scientific approaches \cite{harrison:07}. Such formal
models are however limited to the specific domain and difficult to validate.
Techniques are thus needed that make possible the formal validation, comparison
and extendability of simulation models. As far as we are aware of, the language
presented in this paper, based on modal logic is a first attempt to provide
such a meta model for reasoning about computational models, that has both a
formal semantics as well as the capability to represent realistic concepts.

%
%

\section{Conclusions}\label{sect:conclusions}
Organization concepts and models are increasingly being adopted for the design and specification of
multi-agent systems. The motivations for this model are twofold. In the one hand, the need for a formal,
provable representation of organizations, with their environment, objectives and agents in a way that enables
to analyze their partial contributions to the performance of the organization in a changing environment. On
the other hand, such a model must be realistic enough to incorporate the more `pragmatic' considerations
faced by real organizations.

In this paper we presented an attempt at a formal model for organizational concepts based on
\cite{dignum:omas:chapter:09}, but now including roles and role enacting agents. Although the language itself
seems very rich, because it contains several modal operators, the semantics actually is only a relative small
extension to that of CTL*. We added the set of agent, role tuples to the state transitions to indicate which
role enacting agents are influencing the changes through this transition. This addition allows us to express
not only abilities and achievements of (role enacting) agents, but also their attempts to achieve a state.
Moreover, we can define the same operators for groups of (role enacting) agents. Thus we provide a uniform
framework in which these different concepts can be expressed and combined. Furthermore, we have presented
initial steps towards a complete axiomatization of LAO. The axiomatization presented here gives a good idea of the characteristics of individual
operators and some of the more important relations between operators, but is not intended as a full account. We are currently extending LAO to include the remainder requirements listed in section \ref{sect:intro}.

The current model is based on the notions of controllability, \emph{stit}, attempt and initiative. In the
language presented in this paper we did distinguish between the role (or position) in an organization and the
actual agent performing that role. By making this distinction we have a means to compare two organizations with the same structure, but instantiated with agents having different capabilities. However, in this paper we assumed that agents playing a role will always attempt to
fulfill the objectives of that role. In reality agents might make different choices depending on their other
commitments and preferences. In \cite{dastani:03} we already have shown how the agents can play roles in
different ways. This work could be added to the current framework to incorporate this aspect. This also
allows for the introduction of mental notions such as knowledge and belief in the framework. These notions
will become important when analyzing the performance of organizations with a particular population of agents that might have their own goals and are more or less cooperative, selfish, etc.

Finally, we will extend the model to include deontic concepts and their relation to the operational concepts
presented in this paper. Some of this work has already been done and reported in \cite{davide:phd:07,huib:phd:07}. However, combining the deontic notions with a logic that can distinguish between attempts, achievement and non-performance of actions has not been done before. It brings up interesting questions such as can one be obliged to attempt an action? What is the difference with an obligation to perform the action? Can the difference be checked? Related is the question whether a violation of an obligation where an attempt is made to fulfill it is as "bad" as a violation where no attempt is made at all to fulfill it. We hope to answer some of these questions in future work.


\bibliographystyle{plain}
\bibliography{references,ownpapers}

\end{document}